\PassOptionsToPackage{hidelinks}{hyperref}
\documentclass{article}

\usepackage[preprint]{corl_2026} 
\usepackage{comment}
\usepackage{amsmath,amssymb}
\usepackage{graphicx}
\usepackage{booktabs}
\usepackage{subcaption}
\usepackage{microtype}
\usepackage{braket}
\usepackage{float}
\usepackage{algorithm}
\usepackage{algpseudocode}

\newcommand{\StopGrad}{\operatorname{sg}}
\usepackage{tabularx}
\usepackage{array}
\newcolumntype{Y}{>{\raggedright\arraybackslash}X}
\usepackage{placeins}
\usepackage{longtable}
\usepackage{xurl}
\usepackage{wrapfig}

\title{Guided Discovery of New Behaviors \\ using Diffusion Policies}

\author{
  Dian Yu$^{1}$, Sebastian Sanokowski$^{1}$, and Majid Khadiv$^{1}$\\
  $^{1}$Munich Institute of Robotics and Machine Intelligence, Technical University of Munich\\
  \texttt{\{imdian.yu, sebastian.sanokowski, majid.khadiv\}@tum.de}
}

\begin{document}
\maketitle


\begin{abstract}
Diffusion models have become a powerful tool for generative modeling in robotics, with diffusion policies excelling at modeling multimodal action-trajectory distributions. However, when demonstrations are limited, standard sampling often reproduces dominant behaviors while neglecting valid but rare modes, limiting the discovery of novel solutions. Existing approaches, such as guidance methods or combining reinforcement learning (RL) with diffusion, either push samples into infeasible regions or struggle to escape local minima, failing to systematically uncover diverse behaviors.
To address these challenges, we propose a framework that combines Feynman–Kac correctors with a novel guiding potential that systematically guides diffusion policy samples towards promising yet underrepresented samples. These trajectories are refined using sampling-based trajectory optimization and reincorporated into the training set to retrain the diffusion policy. Our method effectively mines and repairs novel trajectories, enabling the systematic discovery of diverse and executable behaviors. We demonstrate the effectiveness of our framework across a range of manipulation environments, consistently discovering new behaviors. A supplementary video is available at: \href{https://youtu.be/T7MUvMA67VM}{\texttt{youtu.be/T7MUvMA67VM}}.
\end{abstract}


\section{Introduction}
Diffusion models \citep{sohl2015deep, ho2020denoising, song2021scorebased} have emerged as one of the most powerful tools for generative modeling in robotics. In particular, diffusion policies \cite{chi2025diffusionpolicy} are proven to be a powerful tool for modeling multimodal action-trajectory distributions. However, under limited demonstrations, standard sampling often reproduces dominant modes while under-representing valid but rare behaviors~\citep{chi2025diffusionpolicy}. This raises the question: \emph{How can we systematically discover new behaviors in diffusion policies?}\\
A common approach is to use guidance methods \cite{ho2022classifier} to increase sample diversity \cite{saha2024edmp, fan2025diffusion, dai2025safeflow}. However, these methods often push samples into low-support regions, resulting in trajectories that are infeasible or non-executable in the environment. Alternatively, methods which combine reinforcement learning (RL) \cite{sutton1998reinforcement} with diffusion samplers \cite{zhang2021path, berner2024optimal, sanokowski2026rethinking}, such as in \cite{ren2025diffusion, celik2025dime, sanokowski2025diffusion}, could be applied to discover new behaviors. Yet, RL and diffusion samplers are prone to getting stuck in local minima \cite{he2025no}, making it difficult to discover diverse behaviors. Even when initialized with a pretrained policy, RL typically refines the existing policy to improve motion optimality rather than exploring new modes, unless explicit reward shaping toward new goals is used \cite{longhini2026behavioralmodediscoveryfinetuning}. 

\begin{figure}[htbp]
    \centering
    \includegraphics[width=0.85\textwidth]{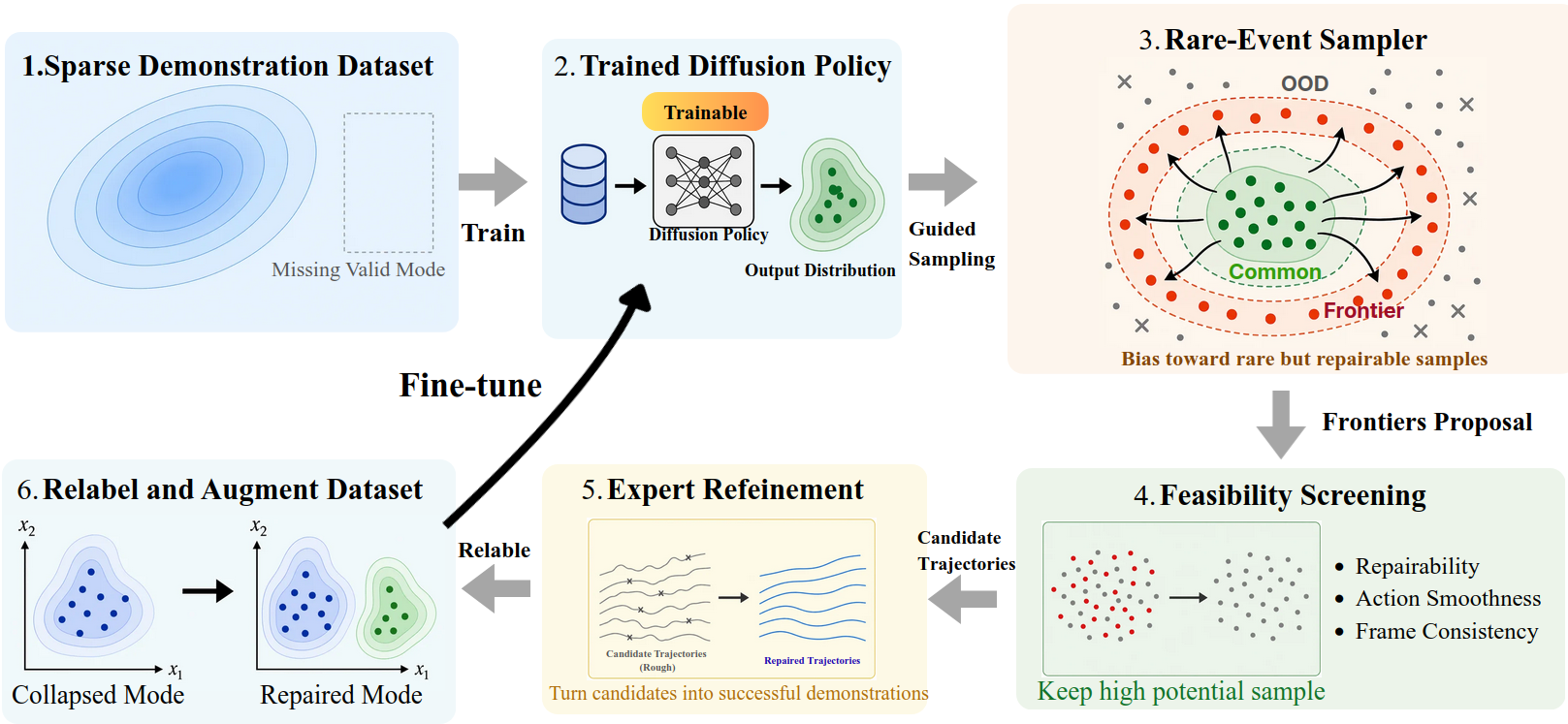}
    \caption{
        Overview of the GDNB bootstrapping loop. 
        A Rare-Event sampler proposes frontier trajectories, which are locally repaired and added back to the dataset for policy fine-tuning.
    }
    \label{fig:framework}
\end{figure}

To address these limitations, we propose our framework \emph{Guided Discovery of New Behaviors} (GDNB), which builds upon guidance methods while mitigating their shortcomings.  GDNB is motivated by the evidence that diffusion models, which are trained in the sparse data regime, can assign low probability to useful behaviors~\citep{gu2023memorization,he2025demystifyingdiffusionpolicies,kadkhodaie2024generalization}. Thus, we systematically guide samples using Feynman--Kac correctors \cite{skreta2025fkc} towards trajectories from the diffusion policy that are both promising and underrepresented samples. These \emph{frontier trajectories} are then refined and corrected using sampling-based trajectory optimization methods and added to the training set to retrain the diffusion policy. Our goal is to systematically mine and repair new trajectories from the diffusion model, enabling the discovery of novel and diverse behaviors.
We evaluate GDNB across diverse manipulation environments and show that it discovers executable behaviors underrepresented by the initial policy. An overview of our framework is presented in Fig.~\ref{fig:framework}.
\subsection{Related Work}
\textbf{Policy improvement and exploration after behavior cloning.}
Several recent methods aim to improve diffusion or generative policies after an imitation learning stage. Diffusion Policy Policy Optimization (DPPO) ~\citep{ren2025diffusion} extends Proximal Policy Optimization (PPO) \cite{schulman2017proximal} to diffusion models and ~\citep{celik2025dime,sanokowski2025diffusion} explain how to apply diffusion models to maximum-entropy reinforcement learning.
DSRL steers a frozen diffusion policy by learning in its latent-noise space~\citep{wagenmaker2025dsrl}; SOE \citep{jin2025soe} constrains robot exploration to a learned action manifold.
However, to discover diverse solutions, these methods need to be combined with skill discovery, novelty search, or reward shaping in order to encourage exploration ~\citep{eysenbach2018diversity,NIPS22RLExplore,Lehman11NoveltySearch, longhini2026behavioralmodediscoveryfinetuning}. 

\textbf{Synthetic demonstrations and local trajectory repair.}
A complementary direction is to synthesize or relabel additional demonstrations. 
MimicGen scales imitation datasets by adapting a small number of human demonstrations to new contexts~\citep{mandlekar2023mimicgen}, while online imitation and dataset-aggregation methods address compounding errors through additional expert or corrective data~\citep{ross2011dagger}. 
Our method instead proposes candidates from the low-support frontier of the learned policy and admits them only after environment rollout and local repair. 
The repair stage builds on sampling-based trajectory optimization \citep{dhedin2026dynaretarget}, leveraging recent advances in zero-order optimization~\citep{kobilarov2012crossentropy,pmlr-v155-pinneri21a,williams2017mppi}. 

\textbf{Guided and rare-event diffusion sampling.}
Inference-time guidance modifies diffusion sampling toward user-specified objectives or constraints~\citep{ho2022classifier}. 
In robotics, guided diffusion or flow-based planners use costs, generated trajectories, or safety constraints to steer samples toward feasible motion~\citep{saha2024edmp,fan2025diffusion,dai2025safeflow}. 

Rare-event diffusion sampling has also been studied in molecular modeling, where several guidance potentials are proposed, that aim at guiding the diffusion model towards rare events~\citep{xie2026enhanced}. 
These methods primarily optimize sampler-side rare-region coverage or unbiased estimator recovery.

\section{Preliminaries}

\label{sec:Problem_Setting}
We consider a sequential decision-making problem with environment time \(t=0,\ldots,T-1\), state \(x_t\in\mathcal X=\mathbb R^{d_x}\), and action \(a_t\in\mathcal A=\mathbb R^{d_a}\). The environment evolves as \(x_{t+1}=f_{\mathrm{dyn}}(x_t,a_t)\), with \(x_0\sim\rho_0\). A rollout is \(\tau=(x_0,a_0,\ldots,x_{T-1},a_{T-1},x_T)\). A sparse task reward \(r_t=r(x_t,a_t,x_{t+1})\), or its rollout aggregate \(R(\tau)\), measures task performance, and \(\operatorname{Success}(\tau)\in\{0,1\}\) denotes task success.
In our setup, we assume access to an offline dataset of demonstrations 
$\mathcal D= \{
        \tau_i
    \}_{i=1}^{N},$
where $\tau_i$ is a successful trajectory. The demonstrations capture valid strategies for solving the task, but typically represent only a sparse subset of all feasible behaviors. In practice, many environments admit multiple qualitatively different solutions that achieve high reward, yet the dataset may over-represent a few dominant modes while under-representing rare but useful behaviors.

Our goal is therefore not only to imitate the demonstrations, but to discover additional successful behaviors while remaining consistent with the environment dynamics and task constraints. Concretely, we seek to learn policies that can interact with the environment in diverse ways while still achieving the task objective.

\subsection{Diffusion Policies}
\label{subsec:diffusion_and_guidance}
Let \(A_t^0\sim p_{\mathrm{data}}(A_t\mid x_t)\), where \(A_t^0=(a_t,\ldots,a_{t+H-1})\in\mathbb R^{H\times d_a}\) is an action chunk of horizon \(H\) conditioned on the current state.
Diffusion policies model multimodal action distributions by learning to reverse a forward noising process that progressively perturbs $A^0_t$ into Gaussian noise. In continuous time, this process is commonly written as a stochastic differential equation (SDE)
\begin{equation}
    dA_t^k
    =
    g(k,A_t^k)\,dk
    +
    b(k)\,dW_k,
    \qquad
    k\in [0,1],
    \label{eq:forward_sde}
\end{equation}
where \(W_k\) is a Wiener process in the action-chunk space \(\mathbb R^{H d_a}\), \(g:[0,1]\times\mathbb R^{H d_a}\to\mathbb R^{H d_a}\) is the forward drift, \(k\in[0,1]\) is the diffusion time step, and \(b:[0,1]\to\mathbb R_+\) is the diffusion coefficient. By using either a variance-preserving or variance-exploding SDE~\citep{song2021scorebased}, the terminal reference distribution is approximately an isotropic Gaussian.

Diffusion models learn to reverse this diffusion process. With \(k\) parameterizing the reverse sampler from noise to data, the reverse-time dynamics are written as
\begin{equation}
    dA_t^k
    =
    \big[
        b(k)^2
        \nabla_{A}
        \log p_k(A_t^k \mid x_t)
        -
        g(k,A_t^k)
    \big]dk
    +
    b(k)\,d\bar W_k,
    \label{eq:reverse_sde}
\end{equation}
where \(\bar W_k\) is reverse-time Brownian motion and \(\nabla_A \log p_k(A_t^k \mid x_t)\) is the conditional score. Since this score is unknown, a neural network \(s_\theta^k(A_t^k,x_t)\) is trained to approximate it.
Then, starting from Gaussian noise, repeated application of Eq.~\eqref{eq:reverse_sde} generates action chunks that approximately follow the demonstrated action distribution. By repeatedly unrolling the diffusion policy in the environment, one obtains the corresponding state trajectories $X_{0:T}$.

For control, we use diffusion policies that generate future action sequences conditioned on the current observation.
At execution time, only the first \(h<H\) actions, \(A_{t,h}=(a_t,\ldots,a_{t+h-1})\), are applied before replanning.
A major advantage of diffusion policies is that they naturally represent multimodal behavior distributions. However, in the low-data regime, sampling is typically dominated by high-density modes from the demonstrations. As a result, rare but valid strategies are often assigned low probability and are rarely generated during ordinary sampling.

\textbf{Guidance toward tilted distributions.}
A common approach to increasing diversity in diffusion model samples is \emph{guidance} \cite{ho2022classifier}. Guidance modifies the reverse diffusion dynamics to bias sampling toward action chunks that maximize a desired property $B(A_t, x_t)$. This can be framed as sampling from a tilted target distribution:
\begin{equation}
    p(A_t) \propto \pi_\theta(A_t|x_t) \exp( \beta B(A_t, x_t)),
    \label{eq:tilted_distr}
\end{equation}
which assigns low probability to regions where either $\pi_\theta(A_t|x_t)$ or $\exp(\beta B(A_t, x_t))$ is small, and high probability where both distributions have significant probability mass.
Standard guidance methods attempt to solve this by adjusting the learned score as $
    \tilde{s}_\theta^k(A_t^k,x_t)
    =
    s_\theta^k(A_t^k,x_t)
    +
    \beta \nabla_A B_k(A_t^k,x_t),$
where $\beta$ controls the guidance strength. However, this approach does not correctly sample from Eq.~\ref{eq:tilted_distr} \citep{skreta2025fkc} and introduces bias, potentially leading to samples with low support under $p(A_t)$. This issue can be addressed using Feynman--Kac correctors (FKCs) \citep{skreta2025fkc}, which provide a principled framework for steering diffusion processes (see Sec.~\ref{sec:guide_rare} for more details).

\subsection{Sampling-Based Trajectory Optimization}
\label{sec:sbto}

Sampling-based trajectory optimization (SBTO) \cite{kurtz2024hydrax} optimizes a candidate action sequence by repeatedly sampling candidate control trajectories and performing a weighted average over rollouts to improve the cost. To handle the curse of dimensionality in long-horizon problems, recent SBTO formulations~\citep{dhedin2026dynaretarget} optimize over a sliding window of trajectories, warm-starting each segment from the solution of the previous segment. This setting is particularly effective when the optimizer tracks an imperfect reference, which is what we need in this work.
In this work, we modify the SBTO in~\citep{dhedin2026dynaretarget} to use it as a repair module
(see Sec.~\ref{subsec:bootstrapping_framework} and App.~\ref{app:sbto_details}).
\section{Method}
\label{sec:method}
Diffusion models trained on robotic data learn a distribution $\pi_\theta(\tau)$ over trajectories, but in the rare data regime tend to converge toward high-probability modes representing ``average'' behaviors \cite{RanaK-RSS25-IMLE,longhini2026behavioralmodediscoveryfinetuning}. 
In robotics this limits the discovery of diverse and novel solutions. Our method addresses this by explicitly targeting \textit{frontier samples}: trajectories with low local support under the current policy-induced distribution \(\pi_{\theta_r}(\tau)\), but not so far from support that they become unrecoverable. Operationally, sampler-side rarity is measured by calibrated denoiser-score energy, while evaluation-side rarity uses the distance-to-measure (DTM) percentile \(u\) defined in Sec.~\ref{subsec:exp2_benchmark_discovery}.
Rare samples serve a dual purpose. First, they act as \textit{out-of-distribution probes}, preventing mode collapse and ensuring the policy explores diverse behaviors across the support of $\pi_\theta(\tau)$. Second, they often represent \textit{novel or edge-case behaviors} that the model has not yet fully learned. By mining these samples, we force exploration of under-represented state-action regions. Without explicitly targeting them, \(\pi_\theta(\tau)\) would continue refining high-probability behaviors rather than discovering new ones. \\
We propose a closed-loop bootstrapping framework that integrates the methods introduced in Sec.~\ref{sec:Problem_Setting} into a cohesive pipeline, as illustrated in Fig.~\ref{fig:framework}. Our iterative process consists of three key steps: (1) we guide a pretrained diffusion model toward rare samples using a Feynman--Kac corrected biasing potential; (2) a sampling-based trajectory optimization algorithm repairs the sampled rare candidates toward physical feasibility and task completion; and (3) corrected trajectories are appended to the dataset to fine-tune the diffusion policy $\pi_\theta$. This creates a \textit{targeted curriculum} \cite{bengio2009curriculum} where the model attempts unusual behaviors, receives corrective feedback, and improves. Compared to random sampling, this is more efficient as it focuses learning on the most informative, under-represented parts of the distribution. Pseudocode is provided in Algorithm~\ref{alg:method_framework_vectorized}.

\subsection{Guidance toward rare events}
\label{sec:guide_rare}
A central component of our framework is the guidance function $B_k(A^k_t, x_t)$, which systematically biases diffusion samples toward low-probability regions. To achieve this goal, we propose to leverage the learned score $s^k_\theta(A^k_t, x_t)$ of the diffusion model as a proxy for identifying such regions.
Intuitively, samples with small score norms tend to lie in stationary, high-probability regions of the distribution. In contrast, large score norms indicate regions where the model drifts, corresponding to low-probability areas. 
To operationalize this, at diffusion time \(k\) we define
\begin{equation}
B_k(A_t^k,x_t)=-\Phi(d_\theta^k(A_t^k,x_t)),\qquad
d_\theta^k(A_t^k,x_t)=\|\tilde s_\theta^k(A_t^k,x_t)\|_2^2 ,
\end{equation}
where \(\tilde s_\theta^k\) is the active score output whitened using mean and variance statistics from ordinary base-policy rollouts at the same diffusion time. Here, \(\Phi\) is inspired by the Lennard--Jones potential \cite{LJDef,sadus2018second}; writing \(x=d/d^*\), its shape and gradient are shown in Fig.~\ref{fig:rare_shell_potential}.
It is given by
\begin{equation}
    \Phi(d)
    =
        \Big (\frac{d^*}{d} \Big)^{p}
        -
        \frac{p}{q} \Big( \frac{d^*}{d} \Big)^{q}
        +
        \Big(\frac{p}{q}-1 \Big ) ,
    \qquad
    p>q>0.
    \label{eq:lennard_jones}
\end{equation}
where \(p\) and \(q\) control shell sharpness, and \(d^*=d_k^*>0\) is the target rare-shell energy at diffusion time \(k\), calibrated from base-sample statistics as detailed in App.~\ref{app:rare_shell_potential}. 

\textbf{Feynman--Kac Correctors for Tilted Target Distributions.}
To guide a diffusion model with a learned score $s_\theta$ toward the tilted target distribution in ~\eqref{eq:tilted_distr}, \citep{skreta2025fkc} propose simulating a batch of particles according to the following SDEs:
{\small
\begin{align}
    dA_t^k
    &=
    \left[
    b_k^2\left(s_\theta^k(A_t^k,x_t)+\beta_k\nabla_A B_k(A_t^k,x_t)\right)
    -g_k(A_t^k)
    \right]dk
    +b_k\,dW_k,\\
    d\ell_k
    &=
    \left[
    \partial_k\beta_k\,B_k(A_t^k,x_t)
    -\langle\nabla_A B_k(A_t^k,x_t),g_k(A_t^k)\rangle
    +\left\langle
    \beta_k\nabla_A B_k(A_t^k,x_t),
    \frac{b_k^2}{2}s_\theta^k(A_t^k,x_t)
    \right\rangle
    \right]dk .
\end{align}
}
We follow the definitions from Sec.~\ref{subsec:diffusion_and_guidance} and \(\ell_k=\log w_k\) is the particle log-weight. After generation, particles are resampled with probabilities \(\operatorname{softmax}(\ell_0)\).    
(See App.~\ref{app:fkc_weights} for more details).

\subsection{Bootstrapping Framework}
\label{subsec:bootstrapping_loop}
\label{subsec:bootstrapping_framework}

\begin{algorithm}[htbp]
\caption{Pseudocode of the Guided Discovery Framework}
\label{alg:method_framework_vectorized}
{\small
\algrenewcommand\algorithmicindent{0.8em}
\algrenewcommand{\algorithmiccomment}[1]{\hfill{\scriptsize$\triangleright$ #1}}
\begin{algorithmic}[1]
\Require initial dataset $\mathcal D_0$, reset distribution $\rho_0$, rounds $R$, number of independent initial states $N$, number of calibration samples $M$, number of rare samples $L$

\State \label{alg:line:train0}
$\pi_{\theta_0} \gets \operatorname{Train}(\mathcal D_0)$
\Comment{current diffusion action policy}

\For{$r = 0, \ldots, R-1$}
    \State \label{alg:line:init_set} $\{x_{0,n}\}_{n=1}^N \sim \rho_0$
    \State $\mathcal D_r^{+} \gets \{\}$

    \For{$n = 1, \ldots, N$}
        \State \label{alg:line:base_trace}
        $\{\tau^{\mathrm{base}}_{m}\}_{m=1}^M \gets \operatorname{BaseRollout}(\pi_{\theta_r}, x_{0,n})$
        \Comment{Sample $M$ independent trajectories for state $x_{0,n}$}

        \State \label{alg:line:calib}
        $\Gamma_{r,n} \gets \operatorname{CalibrateSampler}(\{\tau^{\mathrm{base}}_{m}\}_{m=1}^M)$
        \Comment{compute score statistics over all $M$ trajectories}

        \State \label{alg:line:rare_rollout}
       $\{\tau^{\mathrm{rare}}_{\ell}\}_{\ell=1}^L \gets \operatorname{RareRollout}(\pi_{\theta_r}, \Gamma_{r,n}, x_{0,n})$
       \Comment{sample $L$ rare trajectories for state $x_{0,n}$}

        \State \label{alg:line:repair}
        $\{\tau_\ell^\star\}_{\ell=1}^L
        \gets
        \operatorname{SBTO}
        (\{\tau_\ell^{\mathrm{rare}}\}_{\ell=1}^L, x_{0,n})$
        \Comment{fixed-horizon local repair}

        \State \label{alg:line:newdata}
        $\mathcal D_r^{+}
        \gets
        \mathcal D_r^{+}
        \cup
         \{\tau_\ell^\star:
        \operatorname{Success}(\tau_\ell^\star)=1,\; \ell=1,\ldots,L\}$
        \Comment{insert successful repairs}
    \EndFor

    \State $\mathcal D_{r+1} \gets \mathcal D_r \cup \mathcal D_r^{+}$
    \State \label{alg:line:update}
    $\pi_{\theta_{r+1}} \gets \operatorname{FineTune}(\pi_{\theta_r}, \mathcal D_{r+1})$
\EndFor

\State \Return $\mathcal D_R, \pi_{\theta_R}$
\end{algorithmic}
}
\end{algorithm}

Algorithm~\ref{alg:method_framework_vectorized} summarizes the bootstrapping loop. Starting from one of $N$ randomly sampled initial states $x_{0,n}$, a batch of $M$ calibration samples is obtained from the unguided diffusion model (\(\operatorname{BaseRollout}\)) in order to obtain calibration statistics $\Gamma_{r,n}$. Next, a batch of $L$ candidates is sampled from the rare-event sampler (\(\operatorname{RareRollout}\)) via FKC correction, and these trajectories are further refined using \(\operatorname{SBTO}\). Finally, successful repairs are added to \(\mathcal D_r\), and the diffusion policy is fine-tuned for the next round.

\section{Experiments}

\begin{figure*}[htbp]
\centering
\includegraphics[width=0.90\textwidth]{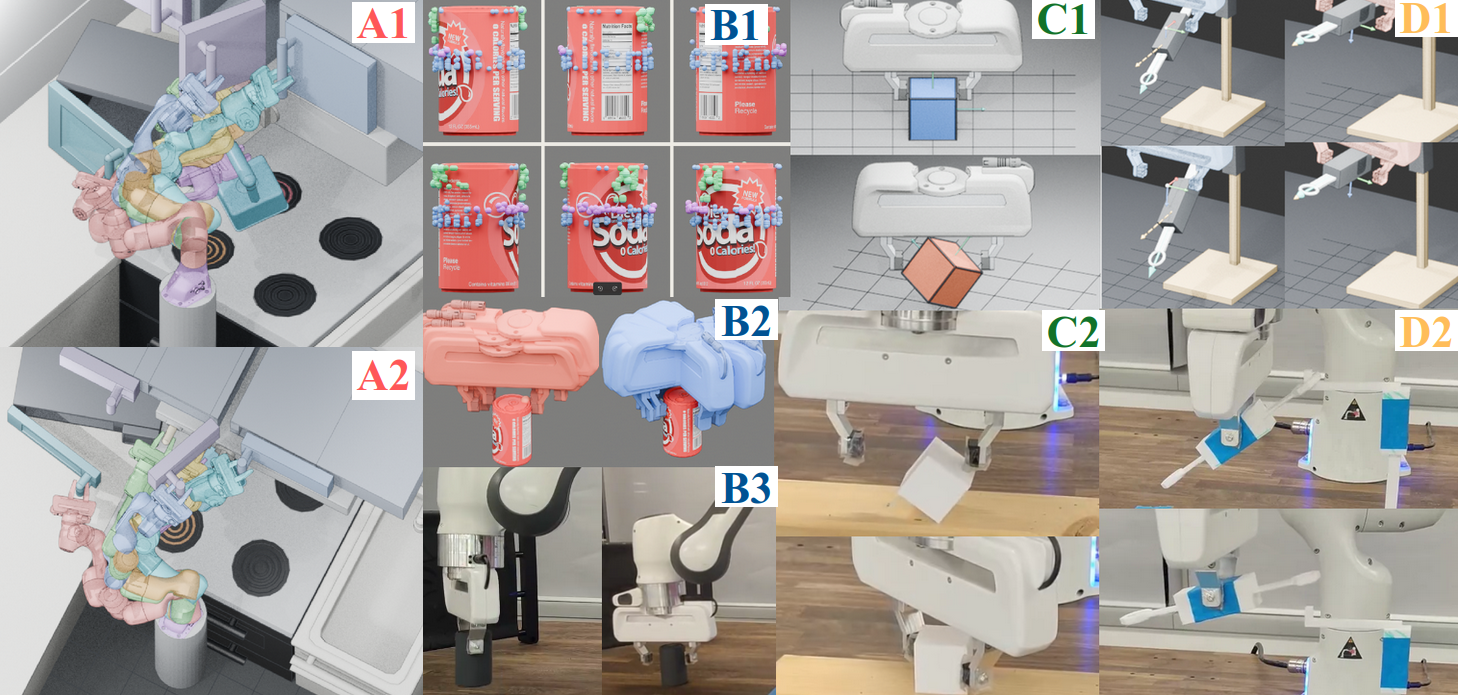}
\caption{
\textbf{Representative GDNB-discovered behaviors and real-world replay.}
(Blue/Cyan) denote common behaviors from the base policy, (orange/red) denote rare behaviors discovered by GDNB. Translucent robot copies indicate temporal rollout snapshots, and colored markers indicate accumulated contact locations. This figure shows key frames and contact/pose summaries.
\textbf{Kitchen} (A1--A2): GDNB discovers rare subtask sequences, including rare successful sequences that complete at least four subtasks; here, colors indicate rollout stages rather than common-vs-rare labels.
\textbf{Can} (B1--B3): B1 shows that GDNB discovers new gripper--can contact locations, B2 compares the corresponding common and rare end-effector grasp poses, and B3 shows a real-robot replay of the rare grasp-pose variant.
\textbf{Lift} (C1--C2): C1 compares a common blue-box pickup with a rare orange/red-box pickup at the lifting moment, and C2 shows the corresponding real-robot replay.
\textbf{ToolHang} (D1--D2): D1 compares a common blue-gripper release motion with a rare orange/red-gripper release orientation, and D2 shows the real-robot replay of the rare release variant.
}
\label{fig:exp2_hero}
\end{figure*}

In our experiments, we aim to evaluate whether our framework enables a diffusion policy to learn new behaviors.
To this end, we first assess in Section~\ref{subsec:turning_double_well} whether our rare event guidance method proposed in Section~\ref{sec:method} combined with SBTO refinement can recover missing modes in a multimodal action distribution starting from a pretrained policy.
In Section~\ref{subsec:exp2_benchmark_discovery}, we first evaluate how our guidance potential performs in sampling rare yet in-distribution samples, comparing it against several baseline methods.
Finally, we apply our full pipeline to challenging robotics manipulation benchmarks in simulation and validate the results with real-world experiments.
\subsection{Mode Recovery in Multimodal Action Distributions}
\label{subsec:turning_double_well}
\begin{wrapfigure}{r}{0.55\linewidth}
    \centering
    \includegraphics[width=0.9\linewidth]{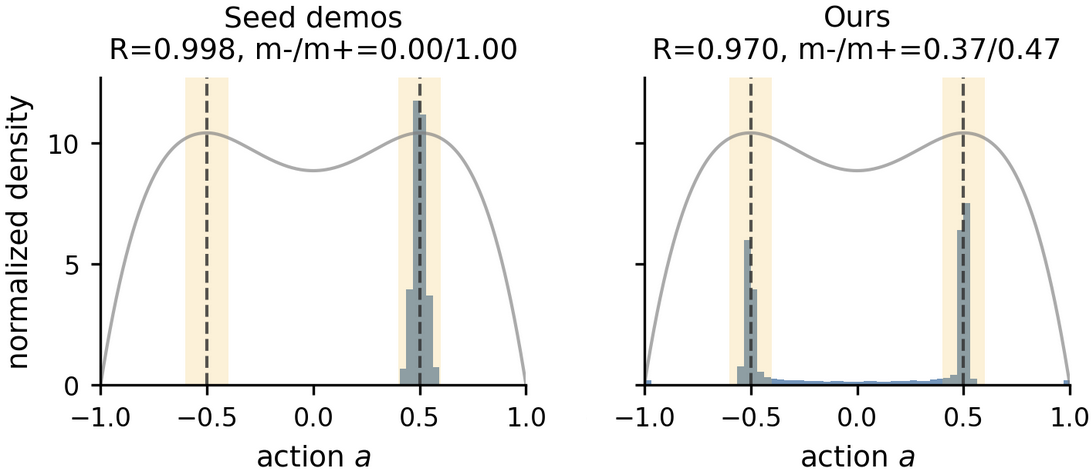}
    \caption{
    \textbf{Mode recovery in \texttt{Multimodal Agent}.}
    Demonstrations cover only the right-turn optimum. GDNB is able to recover the probability mass of the second optimum.
    }
    \label{fig:tdw_triptych}
\end{wrapfigure}

The \texttt{Multimodal Agent} benchmark, introduced in \cite{sanokowski2025diffusion}, is a multimodal RL task designed to evaluate whether diffusion-based RL policies can effectively cover multimodal action distributions in markov decision processes. In this benchmark, an agent proposes movements within a continuous range of possible directions, while the environment dynamics discretize the resulting state into one of eight fixed directional states. The initial demonstrations lie intentionally in a single-mode. Importantly, the reward landscape is bimodal for each state, with reward-equivalent optima at \(a=\pm0.5\), corresponding to \(\pm45^\circ\) turns. For further details, see App.~\ref{app:tdw_details}.
In this work, we repurpose the \texttt{Multimodal Agent} benchmark to test whether, given a collapsed policy that represents only a single mode, GDNB can recover previously undiscovered modes. Our evaluation, illustrated in Figure~\ref{fig:tdw_triptych}, demonstrates that by leveraging the rare-event sampler introduced in Section~\ref{sec:guide_rare} combined with SBTO correction, GDNB successfully recovers the missing modes on this task.

\subsection{Rare Behavior Discovery in Robotic Manipulation}
\label{subsec:exp2_benchmark_discovery}
To systematically evaluate our rare event sampling potential and the discovery of new behavior, GDNB is evaluated on eight robotics manipulation benchmarks from~\citep{chi2025diffusionpolicy}: \textbf{Push-T}, \textbf{Block Pushing}, \textbf{Franka Kitchen}, \textbf{Lift}, \textbf{Can}, \textbf{Square}, \textbf{Transport}, and \textbf{ToolHang}. The suite
covers various object interactions such as planar pushing, object pushing, articulated kitchen manipulation,
grasping, placement, insertion/alignment, bimanual transport, and tool hanging.

\textbf{Action-space rarity sampler performance.}
We first evaluate the rare-event sampler from Sec.~\ref{sec:guide_rare} by analyzing generated action chunks. For each action chunk, we compute a k-nearest-neighbor (kNN) DTM score and convert it to a percentile \(u\in[0,1]\) using held-out unmodified diffusion-policy samples from the same starting condition~\citep{chazal2011geometric} (See App.~\ref{app:action_dtm_percentile}). Equivalently, a larger $u$ indicates that the corresponding samples have lower base policy distribution support. We detect a sample as a frontier sample when $0.90\le u\le0.985$ and report it as \emph{Frontier}. We report the out-of-distribution rate as \emph{OOD} \(=\Pr(u>0.985)\) and the common-sample rate as \emph{Common} \(=\Pr(u<0.90)\). Results are reported in Tab.~\ref{tab:action_dtm_sampler_perf}, where all methods are evaluated with the
same tuning budget. We include two groups of baselines. Existing sampler baselines are UmbrellaDiff, $\Delta G$-Diff, MetaDiff \citep{xie2026enhanced}, and Annealed FKC \cite{skreta2025fkc}. The uncited rows are simple internal baselines (see App.~\ref{app:rare_sampler_variants}):

\begin{wraptable}[15]{r}{0.62\linewidth}
\centering
\scriptsize
\setlength{\tabcolsep}{3pt}
\renewcommand{\arraystretch}{1.03}
\begin{tabular*}{\linewidth}{@{\extracolsep{\fill}}lccc@{}}
\toprule
Sampler
& Frontier $\uparrow$
& OOD $\downarrow$
& Common $\downarrow$ \\
\midrule
\textbf{GDNB (Ours)}
& \(\mathbf{59.05 \pm 2.19}\)
& \(\mathbf{0.00 \pm 0.00}\)
& \(40.95\) \\
\midrule
\multicolumn{4}{@{}l}{\emph{Simple internal baselines}} \\
Direct sampling
& \(4.95 \pm 0.38\)
& \(1.05 \pm 0.18\)
& \(94.00\) \\

Best-of-\(K\) rejection
& \(55.10 \pm 1.95\)
& \(8.10 \pm 0.47\)
& \(36.80\) \\

One-sided action-bank reweighting
& \(45.57 \pm 1.31\)
& \(9.75 \pm 0.53\)
& \(44.68\) \\

Rare-shell Sampling
& \(22.55 \pm 0.77\)
& \(3.90 \pm 0.35\)
& \(73.55\) \\
\midrule
\multicolumn{4}{@{}l}{\emph{Existing sampler baselines}} \\
UmbrellaDiff~\cite{xie2026enhanced}
& \(56.47 \pm 1.99\)
& \(9.88 \pm 0.54\)
& \(33.65\) \\

\(\Delta G\)-Diff~\cite{xie2026enhanced}
& \(21.00 \pm 0.86\)
& \(2.50 \pm 0.28\)
& \(76.50\) \\

MetaDiff~\cite{xie2026enhanced}
& \(16.25 \pm 0.35\)
& \(3.98 \pm 0.15\)
& \(79.77\) \\

Annealed FKC~\cite{skreta2025fkc}
& \(9.45 \pm 0.47\)
& \(1.48 \pm 0.18\)
& \(89.07\) \\
\bottomrule
\end{tabular*}
\caption{
\textbf{Action-space rarity sampler performance.}
Rarity is measured by the base-calibrated kNN-DTM percentile \(u\) on whitened
diffusion action chunks. Values are percentages. Reported \(\pm\) values are standard errors.
}
\label{tab:action_dtm_sampler_perf}
\end{wraptable}

\textbf{Qualitative and real-world evaluation.}
We apply the full GDNB pipeline to manipulation benchmarks, illustrating new behavioral discoveries in Fig.~\ref{fig:exp2_hero}. GDNB's rare rollouts reveal coherent, task-relevant interactions, not mere geometric deviations. The supplementary video includes all simulated rollouts and real-robot replays; Appendix~\ref{app:qualitative_rare_case_descriptions} provides qualitative descriptions for Push-T, Block Pushing, Square, and Transport.
In \textbf{Kitchen}, rarity emerges at the subtask-sequence level: GDNB uncovers uncommon subtask orders, including rare successful sequences completing at least four subtasks (appearing in $\sim1\%$ of the original distribution). Panels A1--A2 show rendered Kitchen rollouts from different viewpoints.
In \textbf{Can}, GDNB discovers a distinct grasp-and-release strategy: the gripper contacts the can at novel locations (B1) and approaches with a different end-effector orientation (B2), with B3 showing the real-robot replay.
In \textbf{Lift}, GDNB finds a flip-then-grasp pickup: the end effector flips the cube into a new orientation before grasping (bottom row of C1), with C2 showing the real-robot replay.
In \textbf{ToolHang}, GDNB discovers a rare mid-air release mode, where the robot releases the tool toward the hook mid-air (right of D1), replayed on the real robot in D2.
These examples demonstrate GDNB's ability to expose structured interactions: subtask ordering, object-local contact, end-effector pose at contact time, object reorientation before grasping, and release timing/orientation. Physical replays confirm executability of selected Can, Lift, and ToolHang motions beyond simulation: after calibrating end-effector--object geometry to match simulation, we replay trajectories open loop on the real robot, recomputing joint-space commands via inverse kinematics to address physical base height and workspace differences.

\textbf{Quantitative evaluation.}
We quantitatively evaluate discovered rollouts using physically interpretable task-related event measures, such as contact geometry, object motion, approach or release pose, handoff geometry, timing, and subtask order. The full list of event measures is provided in App.~\ref{app:rare_event_details}. These events describe the contact geometry, object
motion, approach or release pose, handoff geometry, timing, and subtask order.
We then compare the resulting
sample distributions using Kernel Density Estimation (KDE) \cite{rosenblatt1956remarks}.  We compute two KDE-based metrics. The \emph{Task Feature KDE} is estimated
directly on the scalar event value extracted from the rollout. This metric
measures how the rollout distribution shifts or spreads along a physically
interpretable event dimension. The \emph{Task Rareness KDE} is computed based on a rareness score \(s_a(\tau)\in[0,1]\) (see App.~\ref{app:rare_event_details}), where larger values denote
events that are rarer under the original policy.  Axis definitions, score families, and KDE construction are given in
App.~\ref{app:rare_event_details}; Task-feature distribution-shift diagnostics are shown in
App.~\ref{app:task_feature_kde_overlaps}.
Table~\ref{tab:exp2_four_objective} reports success and selected KDE summary scores. \emph{Feature bal.} balances retained task-feature density and breadth; \emph{DV bal.} balances outside-base density and variance; Memory, Novel gain, and Support denote retained base density, added rare-tail density, and total successful support in rareness space. \(\pm\) denotes standard deviation over the eight benchmarks; (more details in App.~\ref{app:ablation_all_metrics}).

\textbf{Baseline Comparison.}
We compare against five baseline methods that aim to improve or expand a behavior-cloned diffusion policy after initial training: policy-gradient fine-tuning (DPPO~\citep{ren2025diffusion}), modal-level self-improvement (SIME~\citep{jin2025sime}), on-manifold exploration (SOE~\citep{jin2025soe}), synthetic demonstration generation (MimicGen~\citep{mandlekar2023mimicgen}), and latent-noise steering of a frozen diffusion policy (DSRL~\citep{wagenmaker2025dsrl}). All baselines start from the same pretrained diffusion policy or demonstrations, use the same benchmark reset states, and environment success/reward signals. Implementation details are given in App.~\ref{app:ablation_baselines}. We also include three internal framework ablations in the same table: direct rollout of the pretrained checkpoint, GDNB without the rare sampler, and GDNB without SBTO. These rows isolate whether distributional expansion comes from the rare-guided sampler and whether the repair stage is needed to turn rare drafts into reliable executable data.

\begin{table}[htbp]
\centering
\scriptsize
\setlength{\tabcolsep}{4pt}
\renewcommand{\arraystretch}{1.0}
\resizebox{\linewidth}{!}{
\begin{tabular}{@{}l c cc ccc@{}}
\toprule
& & \multicolumn{2}{c}{Task Feature KDE}
& \multicolumn{3}{c}{Task Rareness KDE} \\
\cmidrule(lr){3-4}
\cmidrule(lr){5-7}
Method
& Success $\uparrow$
& Feature bal. $\uparrow$
& DV bal. $\uparrow$
& Memory $\uparrow$
& Novel gain $\uparrow$
& Support $\uparrow$ \\
\midrule

\textbf{GDNB (Ours)}
& $\mathbf{0.883 \pm 0.257}$
& $0.664 \pm 0.189$
& $\mathbf{0.308 \pm 0.131}$
& $0.813 \pm 0.158$
& $\mathbf{0.100 \pm 0.150}$
& $\mathbf{0.582 \pm 0.232}$ \\

DPPO~\citep{ren2025diffusion}
& $0.796 \pm 0.277$
& $0.503 \pm 0.310$
& $0.276 \pm 0.209$
& $0.689 \pm 0.223$
& $0.069 \pm 0.088$
& $0.483 \pm 0.240$ \\

SIME~\citep{jin2025sime}
& $0.752 \pm 0.291$
& $0.660 \pm 0.238$
& $0.262 \pm 0.145$
& $0.923 \pm 0.054$
& $0.033 \pm 0.037$
& $0.555 \pm 0.257$ \\

SOE~\citep{jin2025soe}
& $0.760 \pm 0.272$
& $0.661 \pm 0.228$
& $0.216 \pm 0.127$
& $\mathbf{0.927 \pm 0.088}$
& $0.020 \pm 0.025$
& $0.524 \pm 0.238$ \\

MimicGen~\citep{mandlekar2023mimicgen}
& $0.751 \pm 0.283$
& $0.640 \pm 0.202$
& $0.246 \pm 0.139$
& $0.855 \pm 0.181$
& $0.027 \pm 0.034$
& $0.528 \pm 0.226$ \\

DSRL~\citep{wagenmaker2025dsrl}
& $0.744 \pm 0.283$
& $\mathbf{0.679 \pm 0.215}$
& $0.238 \pm 0.159$
& $0.903 \pm 0.093$
& $0.022 \pm 0.025$
& $0.534 \pm 0.232$ \\

Pretrained checkpoint
& $0.779 \pm 0.141$
& $0.656 \pm 0.223$
& $0.244 \pm 0.159$
& $0.893 \pm 0.056$
& $0.065 \pm 0.053$
& $0.408 \pm 0.081$ \\

w/o rare sampler
& $0.857 \pm 0.104$
& $0.153 \pm 0.084$
& $0.100 \pm 0.034$
& $0.804 \pm 0.057$
& $0.067 \pm 0.054$
& $0.536 \pm 0.072$\\

w/o SBTO
& $0.778 \pm 0.094$
& $0.265 \pm 0.091$
& $0.102 \pm 0.034$
& $0.709 \pm 0.085$
& $0.042 \pm 0.105$
& $0.512 \pm 0.106$ \\

\bottomrule
\end{tabular}
}

{\footnotesize
DV = density--variance.
}
\caption{
\textbf{Policy quality under Task Feature KDE and Task Rareness KDE.}
Task Feature KDE evaluates successful rollout density on task-feature
coordinates. Task Rareness KDE evaluates successful rollout density on
base-calibrated rareness scores. All KDE metrics are scaled to $[0,1]$.
}
\label{tab:exp2_four_objective}
\end{table}

A comparison in Table~\ref{tab:exp2_four_objective} shows that GDNB achieves the highest success rate,
the strongest Task Feature density--variance score, and the largest Task
Rareness novel gain and support. DSRL attains the highest Feature balanced score
and SOE has the highest Memory, but both add less successful rareness-tail
density and achieve lower task success. Overall, GDNB gives the strongest
combined profile across execution, task-feature expansion, and rareness-score
expansion.

\textbf{Continuation across rare-mining rounds.}
The GDNB pipeline can be applied for several rounds of iterations. Thus, we repeat the whole pipeline on the fine-tuned policy on Can, ToolHang, and Transport, where we continue to observe novel behavior without a success-rate drop. On Can, we observe new contact/release points; on ToolHang, we observe new approach/contact variations. Transport mostly preserves the first-round rare behaviors, with smaller
changes in right-hand approach, handoff timing, and contact pose. App.~\ref{app:multiround_continuation}
reports the consecutive-round distribution metrics.

\section{Limitations}
We did not fully characterize the long-term behavior of our GDNB framework. Specifically, it remains unclear whether, after many iterations, the method begins to saturate, either by failing to discover new behaviors or by producing samples via rare guidance that are no longer repairable by SBTO. Second, GDNB requires careful tuning of both the SBTO optimizer and the biasing-potential hyperparameters. Finally, our strategy of adding newly discovered trajectories into the training dataset might not scale well to a large number of GDNB iterations. Thus, it would be useful to add filters that only keep the most important and diverse samples or to combine our method with continual learning \cite{kirkpatrick2017overcoming} methods.
Other interesting directions for future work are to study round-wise saturation and automatic parameter
selection.

\section{Conclusion}
\label{sec:conclusion}
In this work, we present GDNB, a bootstrapping framework for discovering new executable
behaviors in diffusion policies. GDNB uses Feynman--Kac sampling with a biasing potential that systematically guides samples into lower probability regions. A sampling-based optimization algorithm then filters and repairs these samples through rollouts in simulation. Successfully repaired samples are added
to the training set, and the diffusion policy is fine-tuned on the extended dataset. 
Our experiments show that GDNB is able to recover missing modes in a multimodal action space. Across manipulation
benchmarks we observe the emergence of new behaviors. Representative
real-world replays further confirm that the discovered motions are physically
executable.

\clearpage
\bibliography{ref.bib}

@article{chi2025diffusionpolicy,
  title={Diffusion policy: Visuomotor policy learning via action diffusion},
  author={Chi, Cheng and Xu, Zhenjia and Feng, Siyuan and Cousineau, Eric and Du, Yilun and Burchfiel, Benjamin and Tedrake, Russ and Song, Shuran},
  journal={The International Journal of Robotics Research},
  volume={44},
  number={10-11},
  pages={1684--1704},
  year={2025},
  publisher={Sage Publications Sage UK: London, England}
}

@misc{ho2022classifier,
      title={Classifier-Free Diffusion Guidance}, 
      author={Jonathan Ho and Tim Salimans},
      year={2022},
      eprint={2207.12598},
      archivePrefix={arXiv},
      primaryClass={cs.LG},
      url={https://arxiv.org/abs/2207.12598}, 
}

@inproceedings{sohl2015deep,
  title={Deep unsupervised learning using nonequilibrium thermodynamics},
  author={Sohl-Dickstein, Jascha and Weiss, Eric and Maheswaranathan, Niru and Ganguli, Surya},
  booktitle={International conference on machine learning},
  pages={2256--2265},
  year={2015},
  organization={pmlr}
}

@article{ho2020denoising,
  title={Denoising diffusion probabilistic models},
  author={Ho, Jonathan and Jain, Ajay and Abbeel, Pieter},
  journal={Advances in neural information processing systems},
  volume={33},
  pages={6840--6851},
  year={2020}
}

@article{xie2026enhanced,
  title={Enhanced diffusion sampling: Efficient rare event sampling and free energy calculation with diffusion models},
  author={Xie, Yu and Winkler, Ludwig and Sun, Lixin and Lewis, Sarah and Foster, Adam E and Luna, Jos{\'e} Jim{\'e}nez and Hempel, Tim and Gastegger, Michael and Chen, Yaoyi and Zaporozhets, Iryna and others},
  journal={arXiv preprint arXiv:2602.16634},
  year={2026}
}

@article{fan2025diffusion,
  title={Diffusion trajectory-guided policy for long-horizon robot manipulation},
  author={Fan, Shichao and Yang, Quantao and Liu, Yajie and Wu, Kun and Che, Zhengping and Liu, Qingjie and Wan, Min},
  journal={IEEE Robotics and Automation Letters},
  year={2025},
  publisher={IEEE},
  volume       = {10},
  number       = {12},
  pages        = {12788--12795},
  url          ={https://doi.org/10.1109/LRA.2025.3619794},
  doi          = {10.1109/LRA.2025.3619794},
  bibsource    = {dblp computer science bibliography, https://dblp.org}
}

@InProceedings{Barron_2019_CVPR,
author = {Barron, Jonathan T.},
title = {A General and Adaptive Robust Loss Function},
booktitle = {Proceedings of the IEEE/CVF Conference on Computer Vision and Pattern Recognition (CVPR)},
month = {June},
year = {2019}
}

@inproceedings{ren2025diffusion,
  title={Diffusion policy policy optimization},
  author={Ren, Allen and Lidard, Justin and Ankile, Lars and Simeonov, Anthony and Agrawal, Pulkit and Majumdar, Anirudha and Burchfiel, Benjamin and Dai, Hongkai and Simchowitz, Max},
  booktitle={International Conference on Learning Representations},
  year={2025}
}

@inproceedings{bengio2009curriculum,
  title={Curriculum learning},
  author={Bengio, Yoshua and Louradour, J{\'e}r{\^o}me and Collobert, Ronan and Weston, Jason},
  booktitle={Proceedings of the 26th annual international conference on machine learning},
  pages={41--48},
  year={2009}
}

@article{he2025no,
  title={No trick, no treat: Pursuits and challenges towards simulation-free training of neural samplers},
  author={He, Jiajun and Du, Yuanqi and Vargas, Francisco and Zhang, Dinghuai and Padhy, Shreyas and OuYang, RuiKang and Gomes, Carla and Hern{\'a}ndez-Lobato, Jos{\'e} Miguel},
  journal={arXiv preprint arXiv:2502.06685},
  year={2025}
}

@inproceedings{zhang2021path,
title={Path Integral Sampler: A Stochastic Control Approach For Sampling},
author={Qinsheng Zhang and Yongxin Chen},
booktitle={International Conference on Learning Representations},
year={2022},
url={https://openreview.net/forum?id=_uCb2ynRu7Y}
}

@article{berner2024optimal,
title={An optimal control perspective on diffusion-based generative modeling},
author={Julius Berner and Lorenz Richter and Karen Ullrich},
journal={Transactions on Machine Learning Research},
issn={2835-8856},
year={2024},
url={https://openreview.net/forum?id=oYIjw37pTP},
note={}
}

@article{dai2025safeflow,
  title={SafeFlow: Safe Robot Motion Planning with Flow Matching via Control Barrier Functions},
  author={Dai, Xiaobing and Yang, Zewen and Yu, Dian and Liu, Fangzhou and Sadeghian, Hamid and Haddadin, Sami and Hirche, Sandra},
  journal={arXiv preprint arXiv:2504.08661},
  year={2025}
}

@article{sanokowski2025diffusion,
      title={Diffusion-Augmented Markov Decision Processes for Maximum Entropy Reinforcement Learning}, 
    author={Sanokowski, Sebastian and Patil, Kaustubh},
  journal={arXiv preprint arXiv:2512.02019},
  year={2025} 
}

@misc{kurtz2024hydrax,
  title={Hydrax: Sampling-based model predictive control on GPU with JAX and MuJoCo MJX},
  author={Kurtz, Vince},
  year={2024},
  note={https://github.com/vincekurtz/hydrax}
}

@inproceedings{ sanokowski2026rethinking,
title={Rethinking Losses for Diffusion Bridge Samplers},
author={Sebastian Sanokowski and Lukas Gruber and Christoph Bartmann and Sepp Hochreiter and Sebastian Lehner},
booktitle={The Thirty-ninth Annual Conference on Neural Information Processing Systems},
year={2026},
url={https://openreview.net/forum?id=O58KDUfB4x}
}

@article{schulman2017proximal,
  title={Proximal policy optimization algorithms},
  author={Schulman, John and Wolski, Filip and Dhariwal, Prafulla and Radford, Alec and Klimov, Oleg},
  journal={arXiv preprint arXiv:1707.06347},
  year={2017}
}

@article{kirkpatrick2017overcoming,
  title={Overcoming catastrophic forgetting in neural networks},
  author={Kirkpatrick, James and Pascanu, Razvan and Rabinowitz, Neil and Veness, Joel and Desjardins, Guillaume and Rusu, Andrei A and Milan, Kieran and Quan, John and Ramalho, Tiago and Grabska-Barwinska, Agnieszka and others},
  journal={Proceedings of the national academy of sciences},
  volume={114},
  number={13},
  pages={3521--3526},
  year={2017},
  publisher={National Academy of Sciences}
}

@book{sutton1998reinforcement,
  author       = {Richard S. Sutton and
                  Andrew G. Barto},
  title        = {Reinforcement learning - an introduction},
  series       = {Adaptive computation and machine learning},
  publisher    = {{MIT} Press},
  year         = {1998},
  url          = {http://www.incompleteideas.net/book/first/the-book.html},
  isbn         = {978-0-262-19398-6},
  bibsource    = {dblp computer science bibliography, https://dblp.org}
}

@inproceedings{saha2024edmp,
  title={Edmp: Ensemble-of-costs-guided diffusion for motion planning},
  author={Saha, Kallol and Mandadi, Vishal and Reddy, Jayaram and Srikanth, Ajit and Agarwal, Aditya and Sen, Bipasha and Singh, Arun and Krishna, Madhava},
  booktitle={2024 IEEE International Conference on Robotics and Automation (ICRA)},
  pages={10351--10358},
  year={2024},
  organization={IEEE}
}

@inproceedings{ross2011dagger,
  title = 	 {A Reduction of Imitation Learning and Structured Prediction to No-Regret Online Learning},
  author = 	 {Ross, Stephane and Gordon, Geoffrey and Bagnell, Drew},
  booktitle = 	 {Proceedings of the Fourteenth International Conference on Artificial Intelligence and Statistics},
  pages = 	 {627--635},
  year = 	 {2011},
  editor = 	 {Gordon, Geoffrey and Dunson, David and Dudík, Miroslav},
  volume = 	 {15},
  series = 	 {Proceedings of Machine Learning Research},
  address = 	 {Fort Lauderdale, FL, USA},
  month = 	 {11--13 Apr},
  publisher =    {PMLR},
  url = 	 {https://proceedings.mlr.press/v15/ross11a.html},
}

@article{dhedin2026dynaretarget,
  title={DynaRetarget: Dynamically-Feasible Retargeting using Sampling-Based Trajectory Optimization},
  author={Dhedin, Victor and Taouil, Ilyass and Omar, Shafeef and Yu, Dian and Tao, Kun and Dai, Angela and Khadiv, Majid},
  journal={arXiv preprint arXiv:2602.06827},
  year={2026}
}

@inproceedings{skreta2025fkc,
  title = 	 {Feynman-Kac Correctors in Diffusion: Annealing, Guidance, and Product of Experts},
  author =       {Skreta, Marta and Akhound-Sadegh, Tara and Ohanesian, Viktor and Bondesan, Roberto and Aspuru-Guzik, Alan and Doucet, Arnaud and Brekelmans, Rob and Tong, Alexander and Neklyudov, Kirill},
  booktitle = 	 {Proceedings of the 42nd International Conference on Machine Learning},
  pages = 	 {55906--55949},
  year = 	 {2025},
  editor = 	 {Singh, Aarti and Fazel, Maryam and Hsu, Daniel and Lacoste-Julien, Simon and Berkenkamp, Felix and Maharaj, Tegan and Wagstaff, Kiri and Zhu, Jerry},
  volume = 	 {267},
  series = 	 {Proceedings of Machine Learning Research},
  month = 	 {13--19 Jul},
  publisher =    {PMLR},
  url = 	 {https://proceedings.mlr.press/v267/skreta25a.html},
}

@article{gu2023memorization,
title={On Memorization in Diffusion Models},
author={Xiangming Gu and Chao Du and Tianyu Pang and Chongxuan Li and Min Lin and Ye Wang},
journal={Transactions on Machine Learning Research},
issn={2835-8856},
year={2025},
url={https://openreview.net/forum?id=D3DBqvSDbj},
note         = {Accepted by TMLR}
}

@inproceedings{he2025demystifyingdiffusionpolicies,
    title={Demystifying Robot Diffusion Policies: Action Memorization and a Simple Lookup Table Alternative},
    author={Chengyang He and Xu Liu and Gadiel Mark Sznaier Camps and Joseph Bruno and Guillaume Adrien Sartoretti and Mac Schwager},
    booktitle={The Fourteenth International Conference on Learning Representations},
    year={2026},
    url={https://openreview.net/forum?id=PL0tJOfm7I}
}

@inproceedings{kadkhodaie2024generalization,
    title={Generalization in diffusion models arises from geometry-adaptive harmonic representations},
    author={Zahra Kadkhodaie and Florentin Guth and Eero P Simoncelli and St{\'e}phane Mallat},
    booktitle={The Twelfth International Conference on Learning Representations},
    year={2024},
    url={https://openreview.net/forum?id=ANvmVS2Yr0}
}

@inproceedings{song2021scorebased,
    title={Score-Based Generative Modeling through Stochastic Differential Equations},
    author={Yang Song and Jascha Sohl-Dickstein and Diederik P Kingma and Abhishek Kumar and Stefano Ermon and Ben Poole},
    booktitle={International Conference on Learning Representations},
    year={2021},
    url={https://openreview.net/forum?id=PxTIG12RRHS}
}

@inproceedings{REPPO,
title={Relative Entropy Pathwise Policy Optimization},
author={Claas A Voelcker and Axel Brunnbauer and Marcel Hussing and Michal Nauman and Pieter Abbeel and Radu Grosu and Eric Eaton and Amir-massoud Farahmand and Igor Gilitschenski},
booktitle={The Fourteenth International Conference on Learning Representations},
year={2026},
url={https://openreview.net/forum?id=4vmm8mlHkS}
}

@misc{longhini2026behavioralmodediscoveryfinetuning,
      title={Behavioral Mode Discovery for Fine-tuning Multimodal Generative Policies}, 
      author={Alberta Longhini and David Emukpere and Jean-Michel Renders and Seungsu Kim},
      year={2026},
      eprint={2605.11387},
      archivePrefix={arXiv},
      primaryClass={cs.LG},
      url={https://arxiv.org/abs/2605.11387}, 
}

@article{eysenbach2018diversity,
  title={Diversity is all you need: Learning skills without a reward function},
  author={Eysenbach, Benjamin and Gupta, Abhishek and Ibarz, Julian and Levine, Sergey},
  journal={arXiv preprint arXiv:1802.06070},
  year={2018}
}

@inproceedings{NIPS22RLExplore,
 author = {Devidze, Rati and Kamalaruban, Parameswaran and Singla, Adish},
 booktitle = {Advances in Neural Information Processing Systems},
 editor = {S. Koyejo and S. Mohamed and A. Agarwal and D. Belgrave and K. Cho and A. Oh},
 pages = {5829--5842},
 publisher = {Curran Associates, Inc.},
 title = {Exploration-Guided Reward Shaping for Reinforcement Learning under Sparse Rewards},
 url = {https://proceedings.neurips.cc/paper_files/paper/2022/file/266c0f191b04cbbbe529016d0edc847e-Paper-Conference.pdf},
 volume = {35},
 year = {2022}
}

@Inbook{Lehman11NoveltySearch,
author="Lehman, Joel
and Stanley, Kenneth O.",
editor="Riolo, Rick
and Vladislavleva, Ekaterina
and Moore, Jason H.",
title="Novelty Search and the Problem with Objectives",
bookTitle="Genetic Programming Theory and Practice IX",
year="2011",
publisher="Springer New York",
address="New York, NY",
pages="37--56",
isbn="978-1-4614-1770-5",
doi="10.1007/978-1-4614-1770-5_3",
url="https://doi.org/10.1007/978-1-4614-1770-5_3"
}

@INPROCEEDINGS{RanaK-RSS25-IMLE, 
    AUTHOR    = {Krishan Rana AND Robert Lee AND David Pershouse AND Niko Suenderhauf}, 
    TITLE     = {{IMLE Policy: Fast and Sample Efficient Visuomotor Policy Learning via Implicit Maximum Likelihood Estimation}}, 
    BOOKTITLE = {Proceedings of Robotics: Science and Systems}, 
    YEAR      = {2025}, 
    ADDRESS   = {LosAngeles, CA, USA}, 
    MONTH     = {June}, 
    DOI       = {10.15607/RSS.2025.XXI.158} 
}

@article{LJDef,
author = {Mie, Gustav},
title = {Zur kinetischen Theorie der einatomigen Körper},
journal = {Annalen der Physik},
volume = {316},
number = {8},
pages = {657-697},
doi = {https://doi.org/10.1002/andp.19033160802},
url = {https://onlinelibrary.wiley.com/doi/abs/10.1002/andp.19033160802},
eprint = {https://onlinelibrary.wiley.com/doi/pdf/10.1002/andp.19033160802},
year = {1903}
}

@article{sadus2018second,
  title={Second virial coefficient properties of the nm Lennard-Jones/Mie potential},
  author={Sadus, Richard J},
  journal={The Journal of Chemical Physics},
  volume={149},
  number={7},
  year={2018},
  publisher={AIP Publishing},
  doi     = {10.1063/1.5041320}
}

@InProceedings{pmlr-v155-pinneri21a,
  title = 	 {Sample-efficient Cross-Entropy Method for Real-time Planning},
  author =       {Pinneri, Cristina and Sawant, Shambhuraj and Blaes, Sebastian and Achterhold, Jan and Stueckler, Joerg and Rolinek, Michal and Martius, Georg},
  booktitle = 	 {Proceedings of the 2020 Conference on Robot Learning},
  pages = 	 {1049--1065},
  year = 	 {2021},
  editor = 	 {Kober, Jens and Ramos, Fabio and Tomlin, Claire},
  volume = 	 {155},
  series = 	 {Proceedings of Machine Learning Research},
  month = 	 {16--18 Nov},
  publisher =    {PMLR},
  url = 	 {https://proceedings.mlr.press/v155/pinneri21a.html}
}

@InProceedings{wagenmaker2025dsrl,
  title = 	 {Steering Your Diffusion Policy with Latent Space Reinforcement Learning},
  author =       {Wagenmaker, Andrew and Zhang, Yunchu and Nakamoto, Mitsuhiko and Park, Seohong and Yagoub, Waleed and Nagabandi, Anusha and Gupta, Abhishek and Levine, Sergey},
  booktitle = 	 {Proceedings of The 9th Conference on Robot Learning},
  pages = 	 {258--282},
  year = 	 {2025},
  editor = 	 {Lim, Joseph and Song, Shuran and Park, Hae-Won},
  volume = 	 {305},
  series = 	 {Proceedings of Machine Learning Research},
  month = 	 {27--30 Sep},
  publisher =    {PMLR},
  url = 	 {https://proceedings.mlr.press/v305/wagenmaker25a.html}
}

@article{jin2025soe,
  title={SOE: Sample-Efficient Robot Policy Self-Improvement via On-Manifold Exploration},
  author={Jin, Yang and Lv, Jun and Xue, Han and Chen, Wendi and Wen, Chuan and Lu, Cewu},
  journal={arXiv preprint arXiv:2509.19292},
  year={2025}
}

@InProceedings{celik2025dime,
  title = 	 {{DIME}: Diffusion-Based Maximum Entropy Reinforcement Learning},
  author =       {Celik, Onur and Li, Zechu and Blessing, Denis and Li, Ge and Palenicek, Daniel and Peters, Jan and Chalvatzaki, Georgia and Neumann, Gerhard},
  booktitle = 	 {Proceedings of the 42nd International Conference on Machine Learning},
  pages = 	 {6958--6977},
  year = 	 {2025},
  editor = 	 {Singh, Aarti and Fazel, Maryam and Hsu, Daniel and Lacoste-Julien, Simon and Berkenkamp, Felix and Maharaj, Tegan and Wagstaff, Kiri and Zhu, Jerry},
  volume = 	 {267},
  series = 	 {Proceedings of Machine Learning Research},
  month = 	 {13--19 Jul},
  publisher =    {PMLR},
  url = 	 {https://proceedings.mlr.press/v267/celik25a.html},
}

@InProceedings{mandlekar2023mimicgen,
  title = 	 {MimicGen: A Data Generation System for Scalable Robot Learning using Human Demonstrations},
  author =       {Mandlekar, Ajay and Nasiriany, Soroush and Wen, Bowen and Akinola, Iretiayo and Narang, Yashraj and Fan, Linxi and Zhu, Yuke and Fox, Dieter},
  booktitle = 	 {Proceedings of The 7th Conference on Robot Learning},
  pages = 	 {1820--1864},
  year = 	 {2023},
  editor = 	 {Tan, Jie and Toussaint, Marc and Darvish, Kourosh},
  volume = 	 {229},
  series = 	 {Proceedings of Machine Learning Research},
  month = 	 {06--09 Nov},
  publisher =    {PMLR},
  url = 	 {https://proceedings.mlr.press/v229/mandlekar23a.html}
}

@article{kobilarov2012crossentropy,
  title={Cross-entropy motion planning},
  author={Kobilarov, Marin},
  journal={The International Journal of Robotics Research},
  volume={31},
  number={7},
  pages={855--871},
  year={2012},
  publisher={SAGE Publications Sage UK: London, England}
}

@article{williams2017mppi,
  title={Model predictive path integral control: From theory to parallel computation},
  author={Williams, Grady and Aldrich, Andrew and Theodorou, Evangelos A},
  journal={Journal of Guidance, Control, and Dynamics},
  volume={40},
  number={2},
  pages={344--357},
  year={2017},
  publisher={American Institute of Aeronautics and Astronautics}
}

@inproceedings{jin2025sime,
  title={Sime: Enhancing policy self-improvement with modal-level exploration},
  author={Jin, Yang and Lv, Jun and Yu, Wenye and Fang, Hongjie and Li, Yong-Lu and Lu, Cewu},
  booktitle={2025 IEEE/RSJ International Conference on Intelligent Robots and Systems (IROS)},
  pages={9792--9799},
  year={2025},
  organization={IEEE}
}

@article{chazal2011geometric,
  title={Geometric inference for probability measures},
  author={Chazal, Fr{\'e}d{\'e}ric and Cohen-Steiner, David and M{\'e}rigot, Quentin},
  journal={Foundations of Computational Mathematics},
  volume={11},
  number={6},
  pages={733--751},
  year={2011},
  publisher={Springer}
}

@article{gu2019statistical,
  title={Statistical analysis of nearest neighbor methods for anomaly detection},
  author={Gu, Xiaoyi and Akoglu, Leman and Rinaldo, Alessandro},
  journal={Advances in Neural Information Processing Systems},
  volume={32},
  year={2019}
}

@article{rosenblatt1956remarks,
  title={Remarks on Some Nonparametric Estimates of a Density Function},
  author={Rosenblatt, Murray},
  journal={The Annals of Mathematical Statistics},
  volume={27},
  number={3},
  pages={832--837},
  year={1956},
  publisher={Institute of Mathematical Statistics}
}

\clearpage
\appendix
\onecolumn

\section{Sampler Details}
\label{app:sampler_details}

This appendix gives sampler-side derivations and implementation details omitted
from the main method. In this sampler appendix only, \(t\) denotes diffusion/reverse time; environment time is not used in the sampler derivations. App.~\ref{app:z_reference} explains the
predicted-noise rarity coordinate used by the rare-event sampler.
App.~\ref{app:action_dtm_percentile} defines the action-space DTM percentile
used in Tab.~\ref{tab:action_dtm_sampler_perf} and summarizes the action-bank
ablation variants. App.~\ref{app:rare_shell_potential} records the
complementary rare-shell potential used by the diffusion sampler.
App.~\ref{app:fkc_weights} derives the Euler--Maruyama Feynman--Kac weight
update used in implementation. Except for the post-hoc action-space DTM
coordinate, quantities in this section are defined in denoising coordinates and
do not define a task-space interest region or a rollout-side frontier gate.

\subsection{Gaussian-Reference Scale for Denoiser-Score Energy}
\label{app:z_reference}

This appendix gives the reference calculation behind the denoiser-score energy
\(d_\theta^t\) used by the sampler in Sec.~\ref{sec:guide_rare}. The calculation
is used only to put the sampler-side energy on a stable scale. It is not a claim
that diffusion-policy trajectories are globally Gaussian, nor that the resulting
statistic is a calibrated likelihood, task-space novelty score, or standalone
OOD detector.

Let
\[
    \tilde s_\theta^t(y,c)
    =
    \frac{
        P_{\mathrm{act}}\epsilon_\theta(y,t,c)-\mu_t^s
    }{
        \sigma_t^s+\varepsilon_{\mathrm{std}}
    }
    \in\mathbb R^{d_{\mathrm{eff}}}
\]
be the whitened active denoiser-score proxy at reverse time \(t\), where
\(P_{\mathrm{act}}\) removes padded or inactive denoiser-output coordinates.
The sampler-side calibration bank in Algorithm~\ref{alg:method_framework_vectorized}
provides time-dependent statistics \(\mu_t^s,\sigma_t^s\), fitted from ordinary
policy traces. The energy and standardized coordinate are
\begin{equation}
    d_\theta^t(y,c)
    =
    \|\tilde s_\theta^t(y,c)\|_2^2,
    \qquad
    z_t(y,c)
    =
    \frac{d_\theta^t(y,c)-d_{\mathrm{eff}}}{\sqrt{2d_{\mathrm{eff}}}} .
    \label{eq:app_z_def}
\end{equation}
The whitening is a calibration step for denoiser-score proxy features across
reverse time. It is not a density model.

\paragraph{Bayes denoising identity.}
Under the standard $\epsilon$-prediction objective, the population-optimal full predicted-noise vector satisfies
\begin{equation}
    \epsilon_\theta^{\star,\mathrm{full}}(y,t,c)
    =
    \mathbb E[\epsilon\mid Y_t=y,c]
    =
    -\bar\sigma_t\nabla_y\log p_t(y\mid c),
    \label{eq:app_bayes_full}
\end{equation}
on the coordinates of the Gaussian corruption process. 
Projecting to the active coordinates gives
\begin{equation}
    h_t^\star(y,c)
    =
    P_{\mathrm{act}}\epsilon_\theta^{\star,\mathrm{full}}(y,t,c)
    =
    -\bar\sigma_t
    P_{\mathrm{act}}\nabla_y\log p_t(y\mid c).
    \label{eq:app_bayes_active}
\end{equation}
Thus, the active predicted-noise output is proportional to the active conditional score; after whitening, it is the implementation of the main-text proxy \(\tilde s_\theta^t\), whose squared norm is \(d_\theta^t\).

\paragraph{Reference scale.}
If the active whitened coordinates were locally close to independent standard Gaussian residuals, then \(d_\theta^t\) would have mean \(d_{\mathrm{eff}}\) and variance \(2d_{\mathrm{eff}}\). 
This motivates the standardized coordinate in Eq.~\eqref{eq:app_z_def}. 
In practice, the policy distribution can be sparse, multimodal, and anisotropic, so the global distribution of \(d_\theta^t\) need not be \(\chi^2\). 
We therefore use \(z_t\) only as a proposal-side coordinate for constructing the complementary potential. 
Whether a sampled action becomes data is decided later by forward rollout, local refinement, and acceptance.

\subsection{Action-Space kNN-DTM Percentiles and Action-Bank Variants}
\label{app:action_dtm_percentile}
\label{app:rare_sampler_variants}

Tab.~\ref{tab:action_dtm_sampler_perf} uses an action-space rarity coordinate
computed separately for each starting condition. The coordinate is based on
DTM, a robust distance-like construction for probability
measures~\citep{chazal2011geometric}. Nearest-neighbor DTM scores have also
been studied as anomaly scores with finite-sample guarantees~\citep{gu2019statistical}.
We use this score only as a calibrated local rarity proxy over action chunks,
not as a global likelihood estimator.

For a fixed starting condition \(s\), the base action bank contains action
chunks in \(\mathbb R^{N_{\mathrm{bank}}\times H\times d_a}\), which we flatten to \(x_i\in\mathbb R^{H d_a}\). Whitening is fitted per starting condition
using train-split robust statistics: median centering and MAD scaling, with a
standard-deviation fallback for nearly constant coordinates. We use a \(70\%\)
train split as the reference bank \(\mathcal R_s\) and the remaining split for
calibration. For a whitened query \(\tilde x\), the empirical kNN-DTM score is
\[
    d_k(\tilde x;\mathcal R_s)
    =
    \left[
    \frac{1}{k}
    \sum_{i=1}^{k}
    \left\|
        \tilde x-\operatorname{NN}_i(\tilde x;\mathcal R_s)
    \right\|_2^2
    \right]^{1/2},
    \qquad k=10 .
\]
The held-out calibration split converts this score into a percentile
\[
    u(x)=\widehat F_{\mathrm{cal},s}
    \big(d_k(\tilde x;\mathcal R_s)\big).
\]
Thus, larger \(u\) means that the candidate action chunk lies farther from the
direct-policy action bank for the same starting condition. The frontier interval
used in Tab.~\ref{tab:action_dtm_sampler_perf} is \(0.90\le u\le0.985\). The OOD column reports \(\Pr(u>0.985)\).

\paragraph{Action-bank variants.}
The uncited sampler rows in Tab.~\ref{tab:action_dtm_sampler_perf} are internal
action-bank ablations. They operate on an already generated action bank and do
not modify the diffusion reverse dynamics. For completeness, we also include
GDNB's ranking rule in the last row. All values in Tab.~\ref{tab:action_dtm_sampler_perf} are aggregated over
multiple experimental runs and multiple independently sampled starting
conditions; the reported means and standard errors are computed over this
multi-start evaluation pool.

\begin{table}[H]
\centering
\scriptsize
\setlength{\tabcolsep}{4pt}
\renewcommand{\arraystretch}{1.08}
\begin{tabularx}{\linewidth}{@{}p{0.23\linewidth}X@{}}
\toprule
Paper row & Mathematical idea \\
\midrule
Best-of-\(K\) rejection
&
Post-hoc best-of-bank selection. At each step, draw \(K=50\) available bank
candidates and keep the one closest to the target band \(0.90\le u\le0.985\). \\

One-sided action-bank reweighting
&
Weighted sampling without replacement with
\(w(u)\propto\exp\{5\,\operatorname{sigmoid}((u-0.90)/0.03)\}\). This reduces
common samples but does not explicitly penalize \(u>0.985\). \\

Rare-shell Sampling
&
Weighted sampling without replacement with an LJ-like shell weight
\(w(u)\propto\exp[-3\,V_{\mathrm{LJ}}(u;0.975)]\), soft-capped at \(10\). This
is a soft preference, not a hard frontier filter. \\

GDNB row
&
Frontier-first ranking: candidates inside \(0.90\le u\le0.985\) are prioritized,
common candidates are considered next, and \(u>0.985\) OOD candidates are placed
last. The LJ shell value is used as a secondary score. \\
\bottomrule
\end{tabularx}
\caption{
Action-bank sampler variants described with paper-facing names only.
}
\label{tab:app_action_bank_variant_mapping}
\end{table}

The key distinction is that Rare-shell Sampling and GDNB both use an LJ-shaped
rarity preference, but only GDNB adds the hard frontier-first/OOD-last ordering.
This ordering explains why the GDNB row obtains high frontier coverage while
keeping the OOD rate at zero in Tab.~\ref{tab:action_dtm_sampler_perf}. Rare-shell
Sampling targets a higher shell percentile, \(u_\star=0.975\), but because it is
only stochastic weighted sampling, it can still select many common samples and
some OOD samples.

\subsection{Complementary Rare-Shell Potential}
\label{app:rare_shell_potential}

The main method uses a bounded rare-shell potential in denoiser-score energy. 
Given a target standardized level \(z_\star\), the corresponding target
rare-shell energy in the main-text notation is
\begin{equation}
    d_t^\star
    =
    d_{\mathrm{eff}}
    +
    \sqrt{2d_{\mathrm{eff}}}\,z_\star .
    \label{eq:app_target_energy}
\end{equation}
Dropping the conditioning variable \(c\) for readability, use the calibrated
energy \(d_\theta^t(y,c)\) from Eq.~\eqref{eq:app_z_def}. We define the
normalized score-energy coordinate
\begin{equation}
    x_t(y,c)
    =
    \frac{
        d_\theta^t(y,c)+\varepsilon_E
    }{
        d_t^\star+\varepsilon_E
    } .
    \label{eq:app_shell_radius}
\end{equation}
Here \(\varepsilon_E>0\) is a small numerical stabilizer. The value \(x_t(y,c)\approx 1\) corresponds to the target rare shell. Values
below one have lower calibrated denoiser-score energy than the target, and
values above one have higher calibrated denoiser-score energy. We do not
interpret this ordering as a global likelihood ordering.

The shell cost used in the main experiments is
\begin{equation}
    C_t(y)
    =
    \gamma_t
    \operatorname{cap}_{v_{\max}}
    \left[
        x_t(y)^{-p}
        -
        \frac{p}{q}x_t(y)^{-q}
        +
        \left(\frac{p}{q}-1\right)
    \right],
    \qquad
    p>q>0.
    \label{eq:app_shell_potential}
\end{equation}
The uncapped term has its minimum at \(x_t=1\), so \(C_t\) is minimized at
the target shell. We absorb the guidance strength into this time-dependent
cost. In the main-text notation, the complete bias entering the tilted target is
\(\beta_t B_t(y)=-C_t(y)\).

\paragraph{Activation window.}
The scalar schedule \(\gamma_t\in[0,\infty)\) both scales the bias and activates it only on selected reverse-time intervals. 
This prevents the bias from over-constraining the highest-noise phase and avoids over-editing the final denoising steps. 
All variants below use the same guided Euler--Maruyama proposal and FKC correction; only the scalar cost \(C_t\) changes.

\paragraph{One-sided denoiser-energy potential.}
For ablations, we also use a one-sided complementary cost:
\begin{equation}
    C_t^{\mathrm{one}}(y)
    =
    \gamma_t
    \operatorname{softplus}(-\kappa z_t(y)+\tau_z).
    \label{eq:app_one_sided}
\end{equation}
Here \(\kappa>0\) controls the softplus slope, and \(\tau_z\) is a scalar offset for the standardized coordinate \(z_t\).
This version suppresses common low-\(z_t\) particles but does not explicitly
discourage particles from moving beyond a target shell. In the main-text
notation, the corresponding complete bias satisfies
\(\beta_tB_t^{\mathrm{one}}(y)=-C_t^{\mathrm{one}}(y)\).

\begin{figure*}[t]
    \centering
    \includegraphics[width=0.82\textwidth]{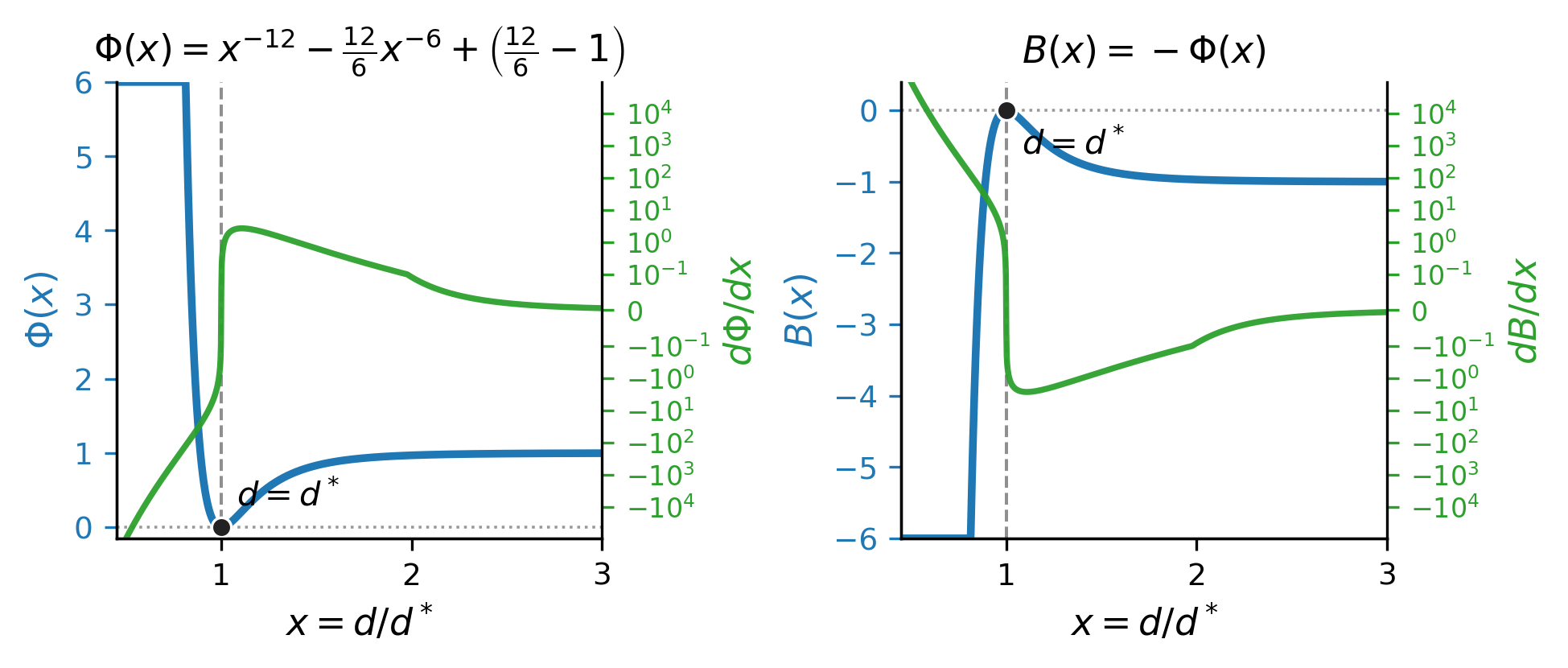}
    \caption{
    \textbf{Lennard-Jones-like rare-shell potential.}
    We visualize the normalized score-energy coordinate \(x=d/d^*\).
    The shell cost \(\Phi(x)=x^{-12}-2x^{-6}+1\) is minimized at \(x=1\),
    corresponding to the target rare shell \(d=d^*\). The reward bias
    \(B(x)=-\Phi(x)\) therefore attracts samples toward this shell instead of
    monotonically pushing them toward increasingly large score norms. Green
    curves show the corresponding derivatives, which determine the guidance
    direction.
    }
    \label{fig:rare_shell_potential}
\end{figure*}
\subsection{Euler--Maruyama Feynman--Kac Weights}
\label{app:fkc_weights}

This appendix derives the discrete weight update used by the rare-event sampler. The derivation is carried out directly on the Euler--Maruyama path space used in implementation. 
This avoids continuous-time divergence, Hessian-trace, or Laplacian estimates.

Let
\[
    Y_k:=Y_{t_k},
    \qquad
    \Delta t_k=t_{k+1}-t_k,
    \qquad
    b_k:=b_{t_k},
    \qquad
    C_k:=C_{t_k}(Y_k),
    \qquad
    \eta_k:=\nabla_yC_{t_k}(Y_k).
\]
Here \(C_t(y):=-\beta_t B_t(y)\) denotes the complete nonnegative shell cost used
by the Euler--Maruyama proposal, including the time-dependent guidance strength.
Thus \(\exp[-C_t(y)]=\exp[\beta_t B_t(y)]\), matching the tilted-target
convention used in Sec.~\ref{sec:guide_rare}.

The unguided Euler--Maruyama target kernel follows the main-text reverse-time drift,
\begin{equation}
    p_k(Y_{k+1}\mid Y_k)
    =
    \mathcal N
    \left(
        Y_{k+1};
        Y_k+
        \left[
            b_k^2s_\theta^{t_k}(Y_k,c)-g(t_k,Y_k)
        \right]\Delta t_k,
        b_k^2\Delta t_k I
    \right).
    \label{eq:app_target_kernel}
\end{equation}

The guided proposal kernel is
\begin{equation}
    q_k^C(Y_{k+1}\mid Y_k)
    =
    \mathcal N
    \left(
        Y_{k+1};
        Y_k+
        \left[
            b_k^2s_\theta^{t_k}(Y_k,c)-g(t_k,Y_k)-b_k^2\eta_k
        \right]\Delta t_k,
        b_k^2\Delta t_k I
    \right),
    \label{eq:app_proposal_kernel}
\end{equation}
where the proposal drift is the main-text unguided reverse-time drift plus the cost-gradient correction,
\begin{equation}
    b_k^2s_\theta^{t_k}(Y_k,c)-g(t_k,Y_k)-b_k^2\eta_k .
    \label{eq:app_guided_drift}
\end{equation}

Equivalently, a proposal sample can be written as
\begin{equation}
    Y_{k+1}
    =
    Y_k
    +
    \left[
        b_k^2s_\theta^{t_k}(Y_k,c)-g(t_k,Y_k)-b_k^2\eta_k
    \right]\Delta t_k
    +
    b_k\sqrt{\Delta t_k}\xi_k,
    \qquad
    \xi_k\sim\mathcal N(0,I).
    \label{eq:app_guided_em}
\end{equation}

The unnormalized tilted density at time \(t_k\) contains the factor
\(\exp[-C_{t_k}(Y_k)]=\exp[\beta_{t_k}B_{t_k}(Y_k)]\) under the
main-text convention \(C_t=-\beta_tB_t\). The incremental log-weight is therefore
\begin{equation}
    \ell_{k+1}-\ell_k
    =
    -
    (C_{k+1}-C_k)
    +
    \log
    \frac{
        p_k(Y_{k+1}\mid Y_k)
    }{
        q_k^C(Y_{k+1}\mid Y_k)
    },
    \qquad
    \ell_k:=\log w_k .
    \label{eq:app_weight_ratio}
\end{equation}
Since the two kernels are Gaussians with equal covariance, the ratio is explicit. 

Using Eq.~\eqref{eq:app_guided_em},
\begin{equation}
    Y_{k+1}
    -
    \left(
        Y_k+
        \left[
            b_k^2s_\theta^{t_k}(Y_k,c)-g(t_k,Y_k)-b_k^2\eta_k
        \right]\Delta t_k
    \right)
    =
    b_k\sqrt{\Delta t_k}\xi_k,
    \label{eq:app_residual_guided}
\end{equation}
whereas relative to the unguided mean,
\begin{equation}
    Y_{k+1}
    -
    \left(
        Y_k+
        \left[
            b_k^2s_\theta^{t_k}(Y_k,c)-g(t_k,Y_k)
        \right]\Delta t_k
    \right)
    =
    b_k\sqrt{\Delta t_k}\xi_k
    -
    b_k^2\Delta t_k\eta_k .
    \label{eq:app_residual_unguided}
\end{equation}

Substitution into the Gaussian log-density ratio gives
\begin{equation}
    \log
    \frac{
        p_k(Y_{k+1}\mid Y_k)
    }{
        q_k^C(Y_{k+1}\mid Y_k)
    }
    =
    -
    \frac{1}{2}b_k^2\Delta t_k\|\eta_k\|_2^2
    +
    b_k\sqrt{\Delta t_k}
    \langle\eta_k,\xi_k\rangle .
    \label{eq:app_kernel_ratio}
\end{equation}

Combining Eq.~\eqref{eq:app_weight_ratio} and Eq.~\eqref{eq:app_kernel_ratio} yields
\begin{equation}
    \ell_{k+1}
    =
    \ell_k
    -
    (C_{k+1}-C_k)
    -
    \frac{1}{2}b_k^2\Delta t_k\|\eta_k\|_2^2
    +
    b_k\sqrt{\Delta t_k}
    \langle\eta_k,\xi_k\rangle .
    \label{eq:app_fkc_weight}
\end{equation}
This correction is exact for the chosen Euler--Maruyama Gaussian kernels. 
It is not a claim that the continuous-time guided drift alone exactly realizes the tilted marginals.

\paragraph{Resampling.}
Weights are normalized within each particle batch. 
For normalized weights \(\tilde w_i\), we use the Kish effective sample size
\begin{equation}
    \operatorname{ESS}
    =
    \frac{1}{\sum_i \tilde w_i^2}.
    \label{eq:app_ess}
\end{equation}
Sequential Monte Carlo (SMC) resampling is triggered when ESS falls below a fixed threshold during denoising. 
After any intermediate resampling, particle weights are reset uniformly. At the final reverse step, the returned particle is selected according to
\begin{equation}
    \Pr(i)
    =
    \frac{\exp(\ell_{0,i})}{\sum_j \exp(\ell_{0,j})}
    =
    \operatorname{softmax}(\ell_0)_i ,
    \label{eq:app_final_resampling}
\end{equation}
which matches the main-text description.

\section{Local Repair and Policy Adaptation}
\label{app:repair_details}

This appendix expands the non-sampling operators in Algorithm~\ref{alg:method_framework_vectorized}. 
These components do not define the rare region. SBTO tests whether each sampled rare draft can be locally refined into a successful trajectory; the adaptation objective specifies how accepted repairs are mixed with rehearsal data.

\subsection{SBTO Implementation: Frontier-Preserving Local Repair}
\label{app:sbto_details}

In GDNB, SBTO denotes the local repair operator applied after rare-candidate
generation. This operator does not define the rare region and is not used as a
planner from scratch. Instead, it receives a diffusion-sampled draft action
sequence and tests whether the draft can be converted into a successful,
executable rollout through bounded local edits. We call this implementation
frontier-preserving because its trust-region and regularization terms are
designed to keep the repaired trajectory close to the sampled draft while
correcting local failures.

\paragraph{Relation to prior SBTO formulations.}
The repair module follows the standard SBTO recipe of sampling candidate
trajectories in a lower-dimensional knot space, rolling them out in simulation,
and updating the sampling distribution using an elite set~\citep{dhedin2026dynaretarget}.
The use case, however, differs from DynaRetarget~\citep{dhedin2026dynaretarget}.
DynaRetarget is designed for dynamically feasible retargeting and uses an
incremental horizon strategy to solve progressively longer problems. In
contrast, our SBTO instance optimizes a fixed-horizon window around a sampled
diffusion-policy draft. Its objective is not to track an external kinematic
reference as closely as possible, but to preserve the draft's frontier behavior
while making the rollout successful and executable.

\paragraph{Knot representation and local trust region.}
Let
\[
    A^0=(a_0^0,\ldots,a_{T-1}^0)\in\mathbb R^{T\times d_a}
\]
be a rare-candidate draft action sequence, with initial state $x_0$ and task
context $c$. We downsample the draft to $N_k$ action knots and reconstruct full
actions by interpolation:
\begin{equation}
    K^0=\Pi(A^0),
    \qquad
    A(K)=\mathcal I(K).
    \label{eq:app_fpsbto_knots}
\end{equation}
The search is restricted to a local trust region around the draft knots:
\begin{equation}
    \mathcal T_m(K^0)
    =
    \left\{
    K:
    \|D_K^{-1}(K_i-K_i^0)\|_\infty
    \le
    \rho_K^{(m)},
    \quad i=1,\ldots,N_k
    \right\},
    \label{eq:app_fpsbto_trust}
\end{equation}
where $D_K$ scales the action channels and $\rho_K^{(m)}$ is the trust-region
radius at Cross-Entropy Method (CEM) iteration $m$. In practice, $\rho_K^{(m)}$
can be warmed up from a small initial radius to the full radius. This makes
early iterations conservative and prevents the optimizer from immediately
leaving the sampled draft.

For each candidate $K$, the simulator executes
\begin{equation}
    \tau(K)=\operatorname{Rollout}(x_0,A(K),c).
    \label{eq:app_fpsbto_rollout}
\end{equation}
We denote the draft rollout by $\tau^0=\tau(K^0)$.

\paragraph{Repair objective.}
SBTO minimizes a rollout-based repair objective
\begin{equation}
\begin{aligned}
    J_{\mathrm{FP\text{-}SBTO}}(K)
    =
    &\;
    J_{\mathrm{task}}(\tau(K))
    +
    \lambda_{\mathrm{track}}J_{\mathrm{track}}(A(K),A^0,\tau(K),\tau^0)
    \\
    &+
    \lambda_{\mathrm{rs}}J_{\mathrm{res\text{-}smooth}}(A(K),A^0)
    +
    \lambda_{\mathrm{knot}}J_{\mathrm{knot}}(K,K^0)
    \\
    &+
    \gamma_{\mathrm{succ}}(\tau(K))
    \left[
        \lambda_{\mathrm{sparse}}J_{\mathrm{sparse}}(A(K),A^0)
        +
        \lambda_{\mathrm{cap}}J_{\mathrm{cap}}(A(K),A^0)
    \right].
\end{aligned}
\label{eq:app_fpsbto_objective}
\end{equation}
The first line encourages task completion while keeping the repaired trajectory
close to the draft. The second line discourages high-frequency residual edits
and large knot displacement. The final line is a success-gated sparse-edit
penalty: once a candidate is close to succeeding, unnecessary edits are
penalized more strongly.

A typical task term is
\begin{equation}
    J_{\mathrm{task}}(\tau)
    =
    -R_{\mathrm{task}}(\tau)
    +
    \lambda_{\mathrm{fail}}\mathbf 1[\operatorname{Success}(\tau)=0],
    \label{eq:app_fpsbto_task}
\end{equation}
where $R_{\mathrm{task}}$ is the environment reward or rollout score.

\paragraph{Draft tracking and residual smoothness.}
The tracking term collects task-specific local preservation costs, such as
action tracking, end-effector tracking, gripper timing, action-bound penalties,
or task-visible descriptor preservation:
\begin{equation}
    J_{\mathrm{track}}
    =
    J_{\mathrm{act}}
    +
    J_{\mathrm{eef}}
    +
    J_{\mathrm{grip}}
    +
    J_{\mathrm{bound}}
    +
    J_{\mathrm{desc}} .
    \label{eq:app_fpsbto_track}
\end{equation}
Not all terms are used in every task backend; Eq.~\eqref{eq:app_fpsbto_track}
summarizes the common interface.

Let the residual action edit be
\[
    \delta a_t=a_t-a_t^0 .
\]
We penalize high-frequency residual edits rather than over-smoothing the whole
candidate action sequence:
\begin{equation}
    J_{\mathrm{res\text{-}smooth}}(A,A^0)
    =
    \frac{1}{T-2}
    \sum_{t=0}^{T-3}
    \left\|
    D_a^{-1}
    \big(
        \delta a_{t+2}-2\delta a_{t+1}+\delta a_t
    \big)
    \right\|_2^2 .
    \label{eq:app_fpsbto_res_smooth}
\end{equation}
The knot prior is
\begin{equation}
    J_{\mathrm{knot}}(K,K^0)
    =
    \frac{1}{N_k}
    \sum_{i=1}^{N_k}
    \|D_K^{-1}(K_i-K_i^0)\|_2^2 .
    \label{eq:app_fpsbto_knot_prior}
\end{equation}

\paragraph{Frame-sparse Welsch edit penalty.}
A central difference from a purely quadratic edit penalty is the frame-sparse
edit regularizer based on the Welsch/Leclerc robust loss~\citep{Barron_2019_CVPR}.
Define the normalized per-frame edit magnitude
\begin{equation}
    g_t(A,A^0)
    =
    \|D_a^{-1}(a_t-a_t^0)\|_2 .
    \label{eq:app_fpsbto_frame_edit}
\end{equation}
The Welsch penalty is
\begin{equation}
    \rho_W(g;\sigma_{\mathrm{W}})
    =
    1-\exp\left(
        -\frac{g^2}{2\sigma_{\mathrm{W}}^2}
    \right),
    \label{eq:app_fpsbto_welsch}
\end{equation}
and the sparse-edit cost is
\begin{equation}
    J_{\mathrm{sparse}}(A,A^0)
    =
    \frac{1}{T}
    \sum_{t=0}^{T-1}
    \rho_W(g_t(A,A^0);\sigma_{\mathrm{W}}).
    \label{eq:app_fpsbto_sparse}
\end{equation}
Because $\rho_W$ saturates for large $g_t$, this term behaves like a soft count
of edited frames. It discourages repairs that spread small changes across many
frames, while allowing a small number of critical frames to change substantially
when needed for success.

We optionally add a soft cap on the number of edited frames:
\begin{equation}
    N_{\mathrm{edit}}(A,A^0)
    =
    \sum_{t=0}^{T-1}
    \rho_W(g_t(A,A^0);\sigma_{\mathrm{W}}),
    \qquad
    J_{\mathrm{cap}}
    =
    \operatorname{softplus}
    \left(
        \frac{N_{\mathrm{edit}}-N_{\mathrm{cap}}}{\tau_{\mathrm{cap}}}
    \right)^2 .
    \label{eq:app_fpsbto_cap}
\end{equation}

\paragraph{Success-gated edit regularization.}
The sparse-edit regularizer is most useful after the optimizer has found a
nearly successful candidate. We therefore gate it by a smooth success score:
\begin{equation}
    \gamma_{\mathrm{succ}}(\tau)
    =
    \operatorname{sigmoid}
    \left(
        \frac{R_{\max}(\tau)-R_{\mathrm{gate}}}{\tau_{\mathrm{gate}}}
    \right),
    \label{eq:app_fpsbto_success_gate}
\end{equation}
where $R_{\max}(\tau)$ is the maximum rollout reward or task progress score.
When a candidate is far from success, CEM mainly optimizes the task objective.
Once the candidate becomes close to success, the Welsch sparse-edit and edit-cap
terms discourage unnecessary deformation of the draft.

\paragraph{CEM update and fallback.}
At CEM iteration $m$, SBTO samples knot candidates from a Gaussian
distribution, projects them into the trust region in Eq.~\eqref{eq:app_fpsbto_trust},
rolls them out, and selects the elite set $\mathcal E_m$ with lowest
$J_{\mathrm{FP\text{-}SBTO}}$. The sampling distribution is updated by
\begin{equation}
    \mu_{m+1}
    =
    (1-\alpha_\mu)\mu_m
    +
    \alpha_\mu
    \frac{1}{|\mathcal E_m|}
    \sum_{K\in\mathcal E_m}
    \operatorname{vec}(K),
    \label{eq:app_fpsbto_cem_mean}
\end{equation}
\begin{equation}
    \Sigma_{m+1}
    =
    (1-\alpha_\Sigma)\Sigma_m
    +
    \alpha_\Sigma
    \operatorname{Cov}
    \left(
        \{
        \operatorname{vec}(K):K\in\mathcal E_m
        \}
    \right)
    +
    \epsilon_\Sigma I .
    \label{eq:app_fpsbto_cem_cov}
\end{equation}
The best candidate over all CEM iterations is returned. If the optimized
candidate does not improve the repair objective over the draft, the backend
falls back to the draft. Thus optimization is allowed to fail without forcing a
large off-draft motion into the dataset.

\paragraph{Acceptance.}
In the implementation used for the main experiments, insertion into the dataset
uses environment success as the hard acceptance condition:
\begin{equation}
    \operatorname{Accept}(\tau^\star)
    =
    \mathbf 1[
        \operatorname{Success}(\tau^\star)=1
    ] .
    \label{eq:app_fpsbto_accept}
\end{equation}
Locality is enforced before this decision by the trust region, knot prior, tracking terms, Welsch sparse-edit penalty, and fallback. Rarity is not used as a post-repair hard acceptance gate. This separation is intentional: the sampler proposes rare candidates, while SBTO only tests whether each rare-candidate draft can be locally repaired into a successful trajectory.

\subsection{Retention-Regularized Adaptation}
\label{app:retention_adaptation}

This appendix expands the fine-tuning step in Algorithm~\ref{alg:method_framework_vectorized}. 
Let $\mathcal D_r^{+}$ be the accepted repaired trajectories after achieved-outcome relabeling, and let $\mathcal D_{\mathrm{reh}}^{(r)}\subset\mathcal D_r$ be a matched rehearsal set. 
For an observation $o$, action trajectory $A$, condition $c$, diffusion time $t$, and Gaussian noise $\epsilon\sim\mathcal N(0,I)$, define
\begin{equation}
    Y_t
    =
    \sqrt{\bar\alpha_t}A
    +
    \sqrt{1-\bar\alpha_t}\epsilon .
    \label{eq:app_adapt_noising}
\end{equation}

The accepted repairs are trained with the standard denoising loss
\begin{equation}
    \mathcal L_{+}(\theta)
    =
    \mathbb E_{\mathcal D_r^{+},t,\epsilon}
    \left[
        \|
            \epsilon-\epsilon_\theta(Y_t,t,o,c)
        \|_2^2
    \right],
    \label{eq:app_l_plus}
\end{equation}
and rehearsal data are trained with
\begin{equation}
    \mathcal L_{\mathrm{reh}}(\theta)
    =
    \mathbb E_{\mathcal D_{\mathrm{reh}}^{(r)},t,\epsilon}
    \left[
        \|
            \epsilon-\epsilon_\theta(Y_t,t,o,c)
        \|_2^2
    \right].
    \label{eq:app_l_reh}
\end{equation}
When retention regularization is used, the previous checkpoint $\theta_r$ is frozen and the adapted denoiser is anchored on rehearsal states:
\begin{equation}
    \mathcal L_{\mathrm{KD}}(\theta;\theta_r)
    =
    \mathbb E_{\mathcal D_{\mathrm{reh}}^{(r)},t,\epsilon}
    \left[
        \left\|
            \epsilon_\theta(Y_t,t,o,c)
            -
            \StopGrad[
                \epsilon_{\theta_r}(Y_t,t,o,c)
            ]
        \right\|_2^2
    \right].
    \label{eq:app_l_kd}
\end{equation}
The adaptation objective is
\begin{equation}
    \theta_{r+1}
    =
    \arg\min_\theta
    \mathcal L_{+}(\theta)
    +
    \lambda_{\mathrm{reh}}\mathcal L_{\mathrm{reh}}(\theta)
    +
    \lambda_{\mathrm{KD}}\mathcal L_{\mathrm{KD}}(\theta;\theta_r)
    +
    \lambda_{\mathrm{anchor}}\|\theta-\theta_r\|_2^2 .
    \label{eq:app_adaptation_objective}
\end{equation}
The rehearsal and distillation terms reduce drift on previously supported behaviors while the accepted repairs introduce new support.

\section{Multimodal Agent protocol details}
\label{app:tdw_details}

\texttt{Multimodal Agent}~\citep{sanokowski2025diffusion} uses heading
observations
\[
    o_t=[\cos\theta_t,\sin\theta_t]
\]
and scalar actions \(a_t\in[-1,1]\). The reward landscape has two symmetric
optimal action modes at \(a=-0.5\) and \(a=+0.5\). We use this task as a
controlled support-recovery diagnostic: the initial data cover all heading
conditions but contain demonstrations from only one action mode.

\paragraph{Seed distribution.}
The seed demonstrations are deliberately one-sided. Headings are sampled from
the eight supported conditions
\[
    \theta\in\{-180^\circ,-135^\circ,-90^\circ,-45^\circ,
    0^\circ,45^\circ,90^\circ,135^\circ\},
\]
and the observation is the corresponding heading embedding
\([ \cos\theta,\sin\theta]\). Actions are sampled only near the right-turn
optimum \(a=+0.5\), with small Gaussian perturbations and clipping to
\([-1,1]\). Thus the seed dataset is condition-diverse in observations but
single-mode in actions. Across the five random seeds used in our runs, this
corresponds to \(960\) one-sided seed demonstrations.

\paragraph{Toy-task bootstrapping protocol.}
In this toy environment, we replace the robot repair module with a simple
reward-curve selector. Candidate actions are generated by the sampler, ranked by
the native reward curve, and the top \(20\%\) are inserted back into the training
set together with rehearsal samples. The sampler itself does not access the
reward curve. This isolates whether sampling and reinsertion can create support
for a missing reward-equivalent mode, rather than whether a trajectory optimizer
can solve a manipulation task.

\paragraph{Metrics.}
We report average per-step return \(\bar R\) and left/right mode masses
\begin{equation}
    m_-=\Pr(|a+0.5|\le 0.1),
    \qquad
    m_+=\Pr(|a-0.5|\le 0.1).
    \label{eq:app_tdw_mode_masses}
\end{equation}
Since the missing mode is \(a=-0.5\), the primary support-recovery diagnostic is
\(m_-\). We also report mode balance
\begin{equation}
    B_{\mathrm{mode}}
    =
    1-
    \frac{|m_+-m_-|}{m_++m_-+\varepsilon},
    \label{eq:app_tdw_mode_balance}
\end{equation}
which is near zero for a single-mode policy and near one for a balanced
two-mode policy. These quantities are support diagnostics; the main purpose of
this task is qualitative mode-recovery analysis rather than a scalar-return
benchmark.

\begin{figure}[H]
\centering

\begin{subfigure}[t]{0.42\linewidth}
    \centering
    \includegraphics[width=\linewidth]{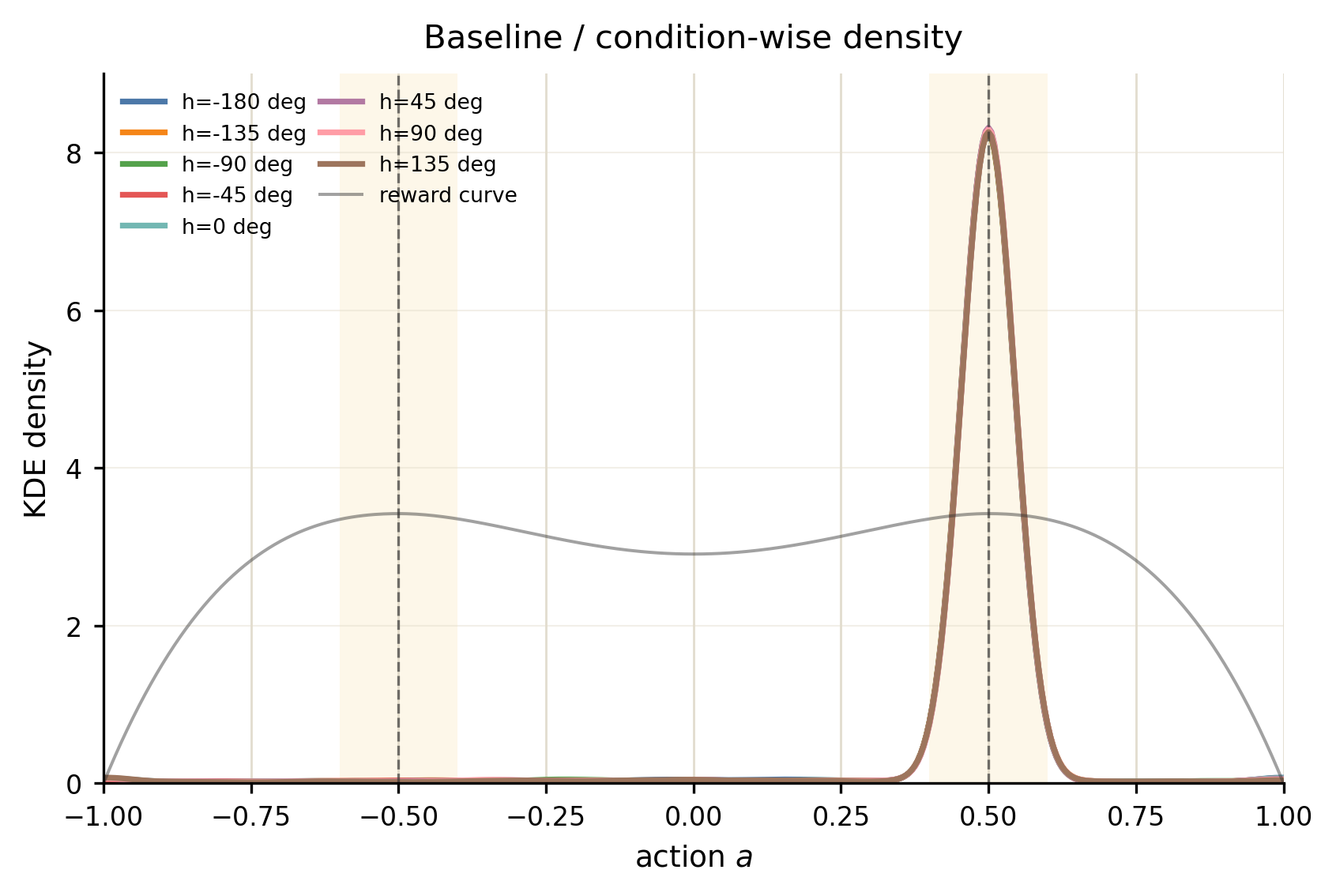}
    \caption{Baseline / seed-trained policy}
    \label{fig:app_tdw_baseline_density}
\end{subfigure}
\hfill
\begin{subfigure}[t]{0.42\linewidth}
    \centering
    \includegraphics[width=\linewidth]{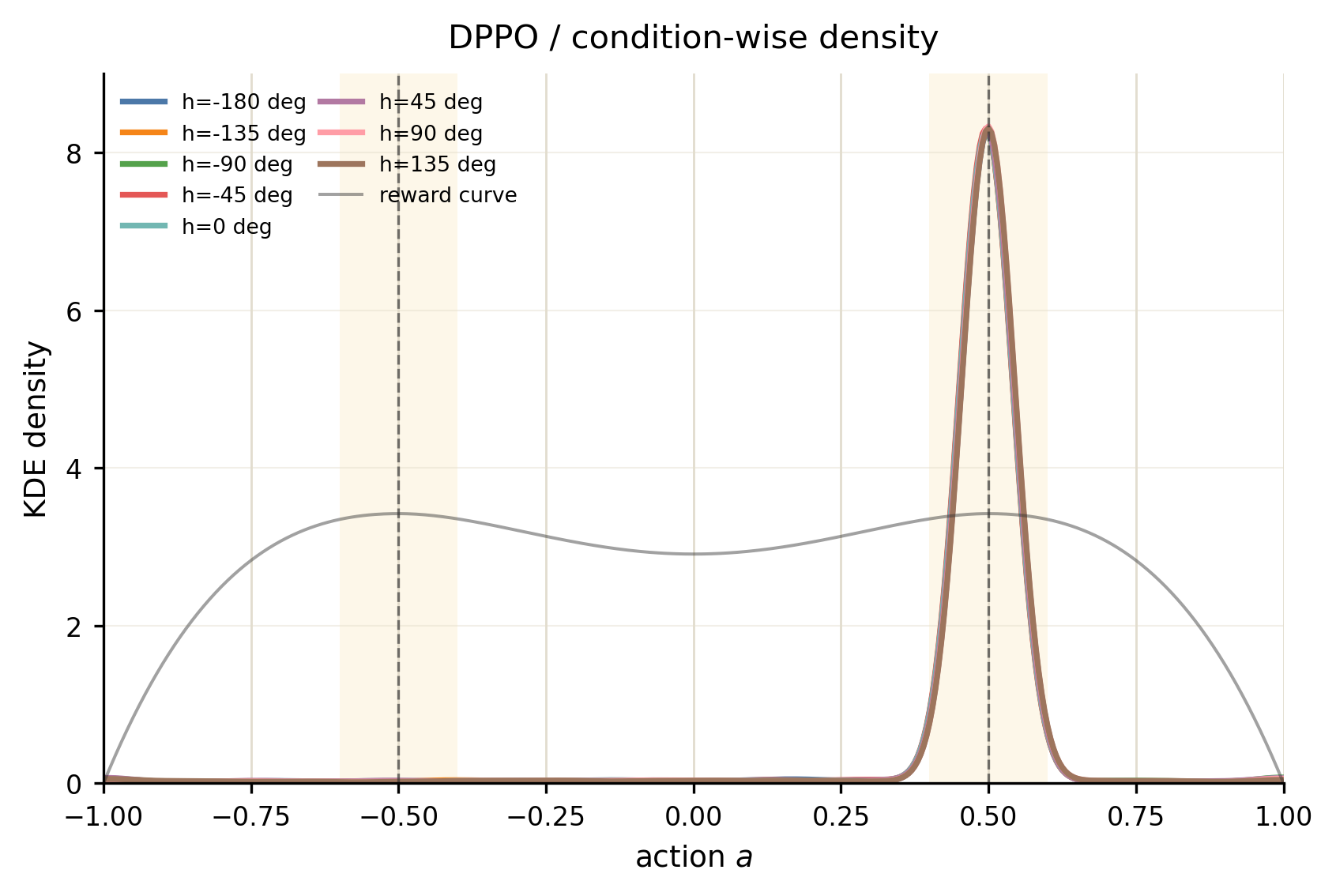}
    \caption{DPPO}
    \label{fig:app_tdw_dppo_density}
\end{subfigure}

\begin{subfigure}[t]{0.42\linewidth}
    \centering
    \includegraphics[width=\linewidth]{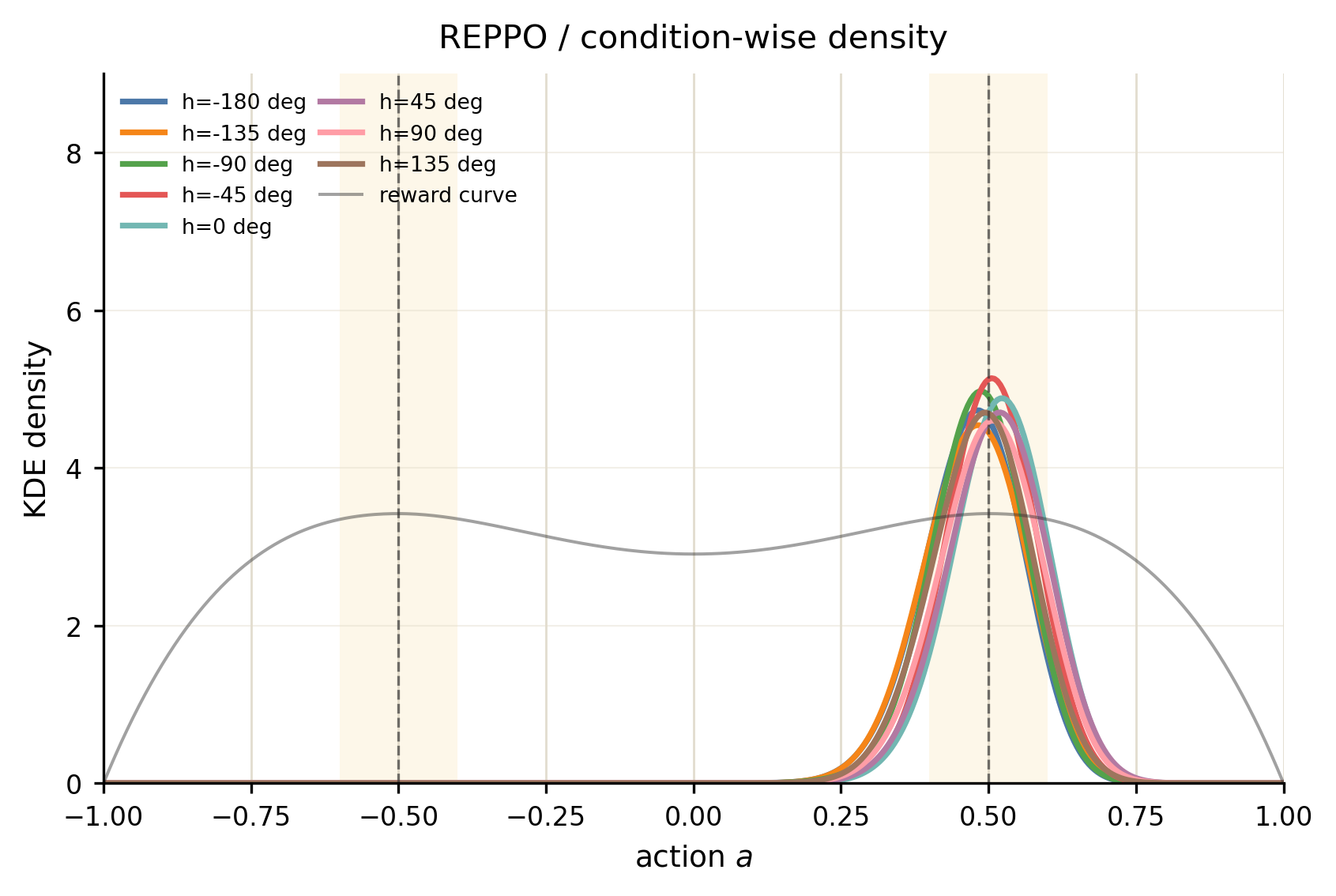}
    \caption{REPPO}
    \label{fig:app_tdw_reppo_density}
\end{subfigure}
\hfill
\begin{subfigure}[t]{0.42\linewidth}
    \centering
    \includegraphics[width=\linewidth]{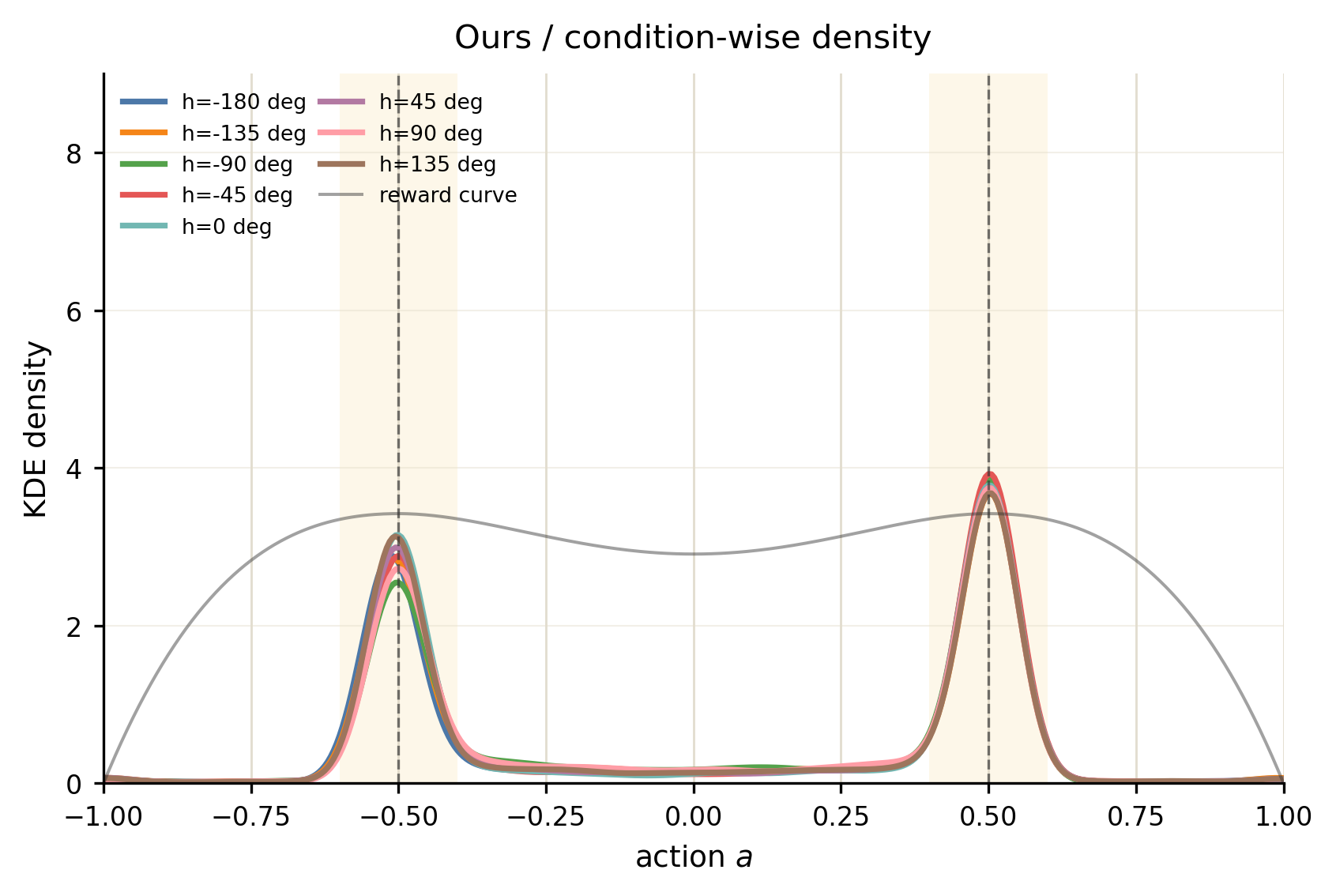}
    \caption{GDNB (Ours)}
    \label{fig:app_tdw_ours_density}
\end{subfigure}

\caption{
\textbf{Condition-wise action densities on \texttt{Multimodal Agent}.}
Each colored curve is the KDE of sampled actions for one heading condition. The
gray curve shows the bimodal reward landscape, and the shaded bands mark the
two reward-equivalent optima at \(a=-0.5\) and \(a=+0.5\). The seed-trained
baseline is concentrated on the right mode. DPPO preserves the same collapse.
REPPO broadens the density around the right well but still does not create a
separate left-mode peak. GDNB recovers visible probability mass near the missing
left mode while retaining the original right mode.
}
\label{fig:app_tdw_condition_wise_density}
\end{figure}

Figure~\ref{fig:app_tdw_condition_wise_density} shows why this benchmark is a
support-recovery test rather than a standard reward-improvement test. Because
both \(a=-0.5\) and \(a=+0.5\) are reward-optimal, a policy can obtain high
return while remaining collapsed onto the seed-supported right mode. The
baseline in Fig.~\ref{fig:app_tdw_baseline_density} and DPPO in
Fig.~\ref{fig:app_tdw_dppo_density} remain sharply concentrated near
\(a=+0.5\). REPPO in Fig.~\ref{fig:app_tdw_reppo_density} increases variance
around the same right-mode region, but it still does not form a distinct density
peak at \(a=-0.5\). In contrast, GDNB in Fig.~\ref{fig:app_tdw_ours_density}
recovers a second peak near the missing left mode while preserving mass near the
original right mode. The qualitative conclusion is therefore support expansion:
native-reward fine-tuning baselines remain essentially single-mode, whereas
rare-guided bootstrapping exposes and reincorporates the absent
reward-equivalent mode.

\section{Task-related Rare Events and KDE Protocol}
\label{app:rare_event_details}

This appendix defines the task-related event measures used in Experiment~2.
These axes are post-hoc evaluation coordinates. They are not used by the
rare-event sampler, the repair optimizer, the fine-tuning loss, or baseline
selection.

\subsection{Qualitative rare-case descriptions for remaining benchmarks}
\label{app:qualitative_rare_case_descriptions}

This subsection complements the main qualitative panels in Fig.~\ref{fig:exp2_hero} by describing the remaining manipulation benchmarks. The descriptions are not additional filtering rules: they are visual interpretations of the same discovered rollout families used for the task-related event measures in Tab.~\ref{tab:exp2_task_axes}.
\begin{figure}[t]
    \centering
    \begin{minipage}[c]{0.42\linewidth}
        \centering
        \includegraphics[width=\linewidth]{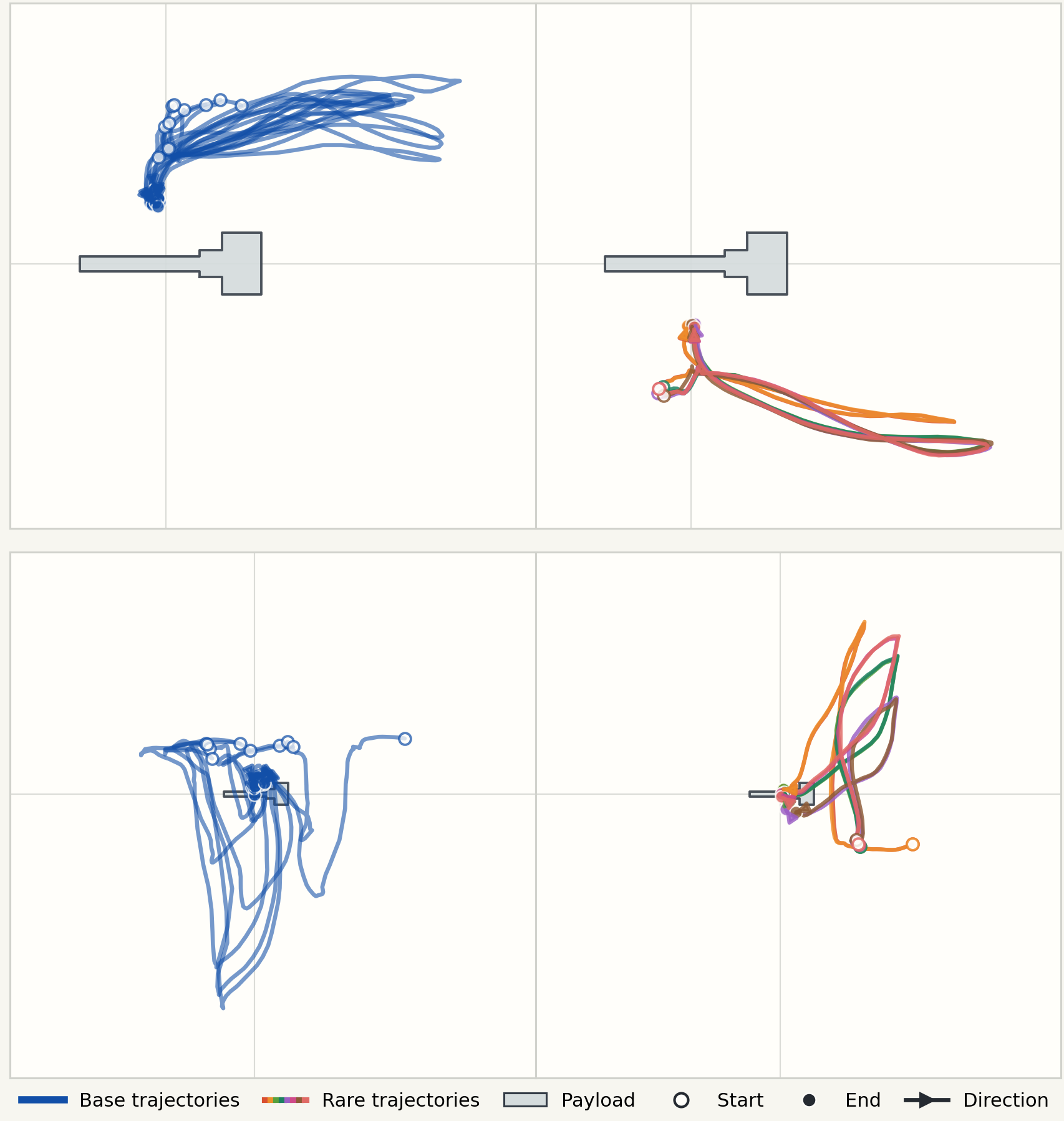}
        \vspace{0.25em}

        {\small\textbf{(a) Transport}}
    \end{minipage}
    \hfill
    \begin{minipage}[c]{0.54\linewidth}
        \centering
        \includegraphics[width=0.90\linewidth]{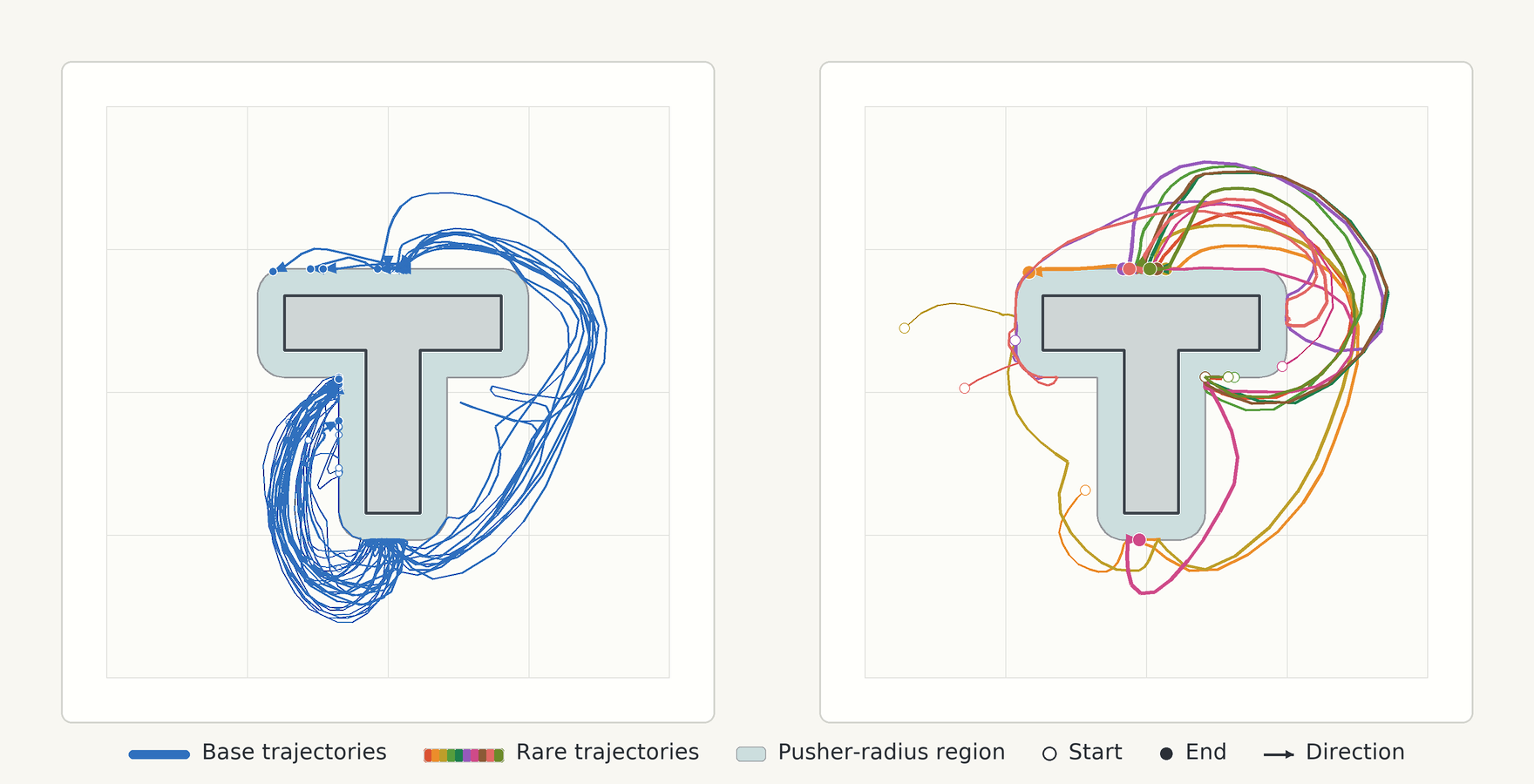}
        \vspace{0.25em}

        {\small\textbf{(b) Push-T}}

        \vspace{0.85em}

        \includegraphics[width=0.90\linewidth]{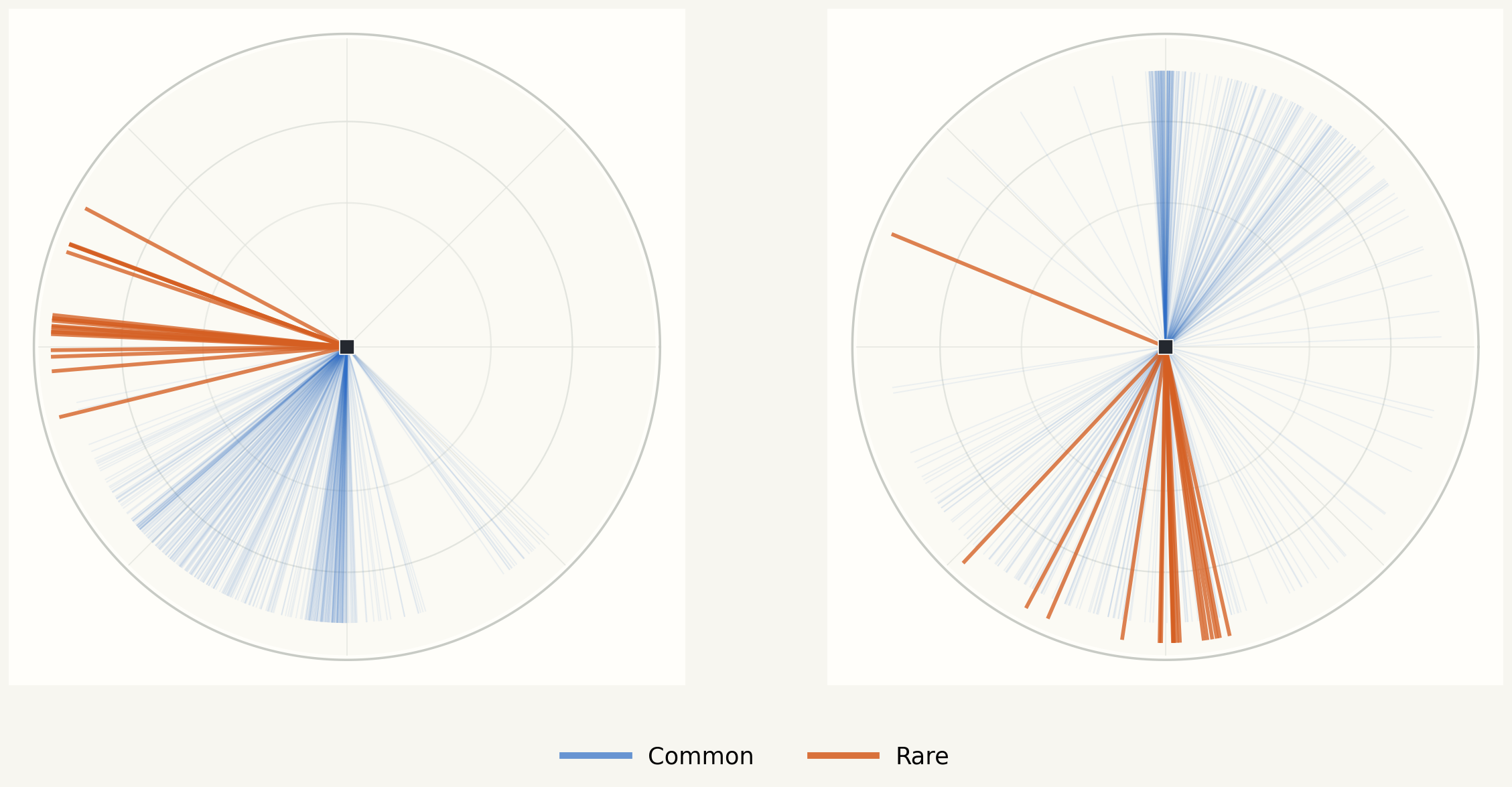}
        \vspace{0.25em}

        {\small\textbf{(c) Block Pushing}}
    \end{minipage}

    \caption{\textbf{Qualitative rare-case diagnostics for remaining benchmarks.}
    (a) End-effector-center trajectories for two phases of Transport, with common rollouts in the left column and rare rollouts in the right column.
    (b) Base and rare pusher-center trajectories around the T-shaped block in Push-T.
    (c) Box-frame angle diagnostics for Block Pushing: first-contact angle on the left and post-contact push direction on the right.}
    \label{fig:app_remaining_rarecases}
\end{figure}

\paragraph{Transport.}
Fig.~\ref{fig:app_remaining_rarecases}(a) visualizes end-effector-center trajectories for two representative phases of the bimanual transport task, with common rollouts shown in the left column and rare rollouts shown in the right column. The first row tracks the left end-effector as it approaches the hammer-shaped payload before lift, ending at the left-clamp proxy. The second row tracks the right end-effector as it approaches the payload while the left end-effector is already holding it, ending at the handoff proxy. Compared with the compact common trajectories, the rare rollouts exhibit different approach corridors and handoff geometries. These differences indicate a task-level behavioral change: GDNB alters the spatial coordination between the two arms during clamp and handoff, rather than merely adding small perturbations to the marginal trajectory of a single end-effector.

\paragraph{Push-T.}
Fig.~\ref{fig:app_remaining_rarecases}(b) visualizes the pusher-center trajectories around the T-shaped object. The base policy concentrates on a narrow family of motions around the commonly used side of the T and repeatedly approaches similar contact neighborhoods. The rare rollouts found by GDNB instead wrap around the object with broad arcs, often reaching the upper shoulder or the opposite side of the T before applying contact. This pattern is physically meaningful because planar pushing is highly sensitive to where the pusher contacts the object: changing the side and angle of contact changes the induced rotation and translation of the T. The rare trajectories therefore correspond to alternative contact sequences for manipulating the same object, rather than to arbitrary geometric perturbations of the base path.

\paragraph{Block Pushing.}
Fig.~\ref{fig:app_remaining_rarecases}(c) shows two angle distributions expressed in the coordinate frame of the box. The left polar plot measures the angle at which the agent first contacts the box, while the right polar plot measures the direction of the subsequent push after contact. Common rollouts occupy the dominant box-frame sectors in blue, whereas rare rollouts concentrate in orange sectors that are only weakly represented by the base policy. The resulting rare behavior is a change in contact mechanics: GDNB discovers side-biased contacts and post-contact pushes from less common box-relative directions. These modes are useful because the same object displacement can be achieved through different contact faces and force directions, and the rare distribution exposes those alternatives explicitly.

\paragraph{Square.}
In Square, the most salient rare behavior is a stable \emph{try--fail--retry} insertion pattern. Instead of executing a single monotone approach and insertion, rare rollouts first attempt to align the object with the square peg or goal, encounter a partial-contact failure such as a misalignment or blocked insertion, retreat or reorient locally, and then attempt the insertion again. This recovery-like sequence is rare under the base policy, which more often follows a direct approach mode. The pattern is physically meaningful because contact-rich insertion tasks frequently require local correction after the first contact; discovering such retry behavior adds a qualitatively different recovery strategy rather than merely adding path noise.

\FloatBarrier

\subsection{Task-feature and rareness-score coordinates}

\FloatBarrier
For each task-related event \(a\), let \(f_a(\tau)\) be the rollout
descriptor vector. We summarize each event with two complementary scalar
coordinates: a physically interpretable task-feature coordinate
\(x_a(\tau)\), and a base-calibrated rareness score \(s_a(\tau)\).

\paragraph{Task Feature KDE.}
Task Feature KDE uses a one-dimensional task-feature coordinate
\begin{equation}
    x_a(\tau)
    =
    \rho_a\,v_a^\top
    W_a\big(f_a(\tau)-\mu_a\big),
    \qquad
    \rho_a\in\{-1,+1\}.
    \label{eq:app_task_feature_axis}
\end{equation}
The whitening statistics $(\mu_a,W_a)$ are fitted on base rollouts. The vector
$v_a$ is the event-axis projection direction, and $\rho_a$ orients the axis so
that larger values correspond to the rarer side of the base rollout bank. KDEs
are computed directly on $x_a(\tau)$.

\paragraph{Task Rareness KDE.}
\begin{table}[h]
\centering
\scriptsize
\setlength{\tabcolsep}{3pt}
\renewcommand{\arraystretch}{0.96}
\begin{tabularx}{\textwidth}{@{}l@{\hspace{4pt}}Y@{\hspace{10pt}}l@{\hspace{4pt}}Y@{}}
\toprule
Benchmark & Rare events
& Benchmark & Rare events \\
\midrule
Push-T
& contact dynamics; block progress; contact timing
& Block Pushing
& block contact mode; contact path; block motion trajectory \\

Franka Kitchen
& attempt order; success order
& Lift
& contact timing; short-horizon box motion; box--end-effector contact point \\

Can
& can-local contact geometry; contact end-effector pose; release end-effector pose
& Square
& end-effector trajectory; object--goal contact dynamics; peg-insert motion \\

Transport
& bimanual pre-approach; gripper contact mode; handoff geometry
& ToolHang
& pre-hook approach; grasp/hook contact; gripper--tool contact mode \\
\bottomrule
\end{tabularx}
\caption{
\textbf{Task-related rare events in manipulation benchmarks:}
Semantic rollout descriptors used for post-hoc rare-behavior evaluation.
Task-feature coordinates are defined below; the full descriptor-to-score-family
mapping is given in Table~\ref{tab:app_event_axis_score_families}.
}
\label{tab:exp2_task_axes}
\end{table}
Task Rareness KDE uses a base-calibrated rareness score
$s_a(\tau)\in[0,1]$. Depending on the event, $s_a$ is computed by kNN-DTM
percentile, robust-distance percentile, raw percentile, or task-interest anchor
percentile. 

For kNN-DTM axes, we use
\begin{equation}
    d_{k,a}(\tau)
    =
    \left[
    \frac{1}{k}
    \sum_{v\in \operatorname{NN}_k(\tilde f_a(\tau);\mathcal V_a)}
    \|\tilde f_a(\tau)-v\|_2^2
    \right]^{1/2},
    \qquad
    s_a(\tau)
    =
    \widehat F_{\mathrm{cal},a}\big(d_{k,a}(\tau)\big),
    \label{eq:app_task_rareness_axis}
\end{equation}

where $\tilde f_a$ is the whitened event feature, $\mathcal V_a$ is the base
reference split, and $\widehat F_{\mathrm{cal},a}$ is the empirical CDF on the
base calibration split.

For Kitchen, attempt-order and success-order are discrete subtask-order descriptors. KDE is applied only after mapping these descriptors to scalar base-calibrated feature/rareness coordinates, so the curves should be interpreted as smoothed empirical summaries rather than densities over raw sequence labels. The separate
true-success-sequence diagnostic is categorical and is analyzed by
sequence frequencies rather than continuous KDE.

\subsection{KDE protocol}

For both KDE views, density-based metrics use success-only rollouts. For method
$m$ and axis $a$, let $\hat p_{m,a}^{F,+}$ be the success-only KDE on the Task
Feature coordinate $x_a$, and let $\hat p_{m,a}^{R,+}$ be the success-only KDE
on the Task Rareness coordinate $s_a$. KDEs are evaluated on a fixed
one-dimensional grid using a Gaussian kernel and are normalized to unit area on
the grid. Metrics are computed per axis, averaged within each benchmark, and
then averaged across the eight benchmarks.

Task Feature KDE reports feature retention, feature breadth, feature balanced,
base-density affinity, outside-base mass, outside-base variance,
outside-base evenness, outside-base diversity, and density--variance balanced.
Task Rareness KDE reports Memory, Novel gain, Support, and Rare recall.
\subsection{Task event measures and score families}
\label{app:event_axis_score_families}

\begingroup
\scriptsize
\setlength{\LTpre}{0.4em}
\setlength{\LTpost}{0.4em}
\renewcommand{\arraystretch}{1.06}
\begin{longtable}{@{}p{0.13\linewidth}p{0.45\linewidth}p{0.33\linewidth}@{}}
\caption{
\textbf{Task event descriptors used by Task Feature KDE and Task Rareness KDE.}
Task Feature KDE uses the listed event descriptor as its task-feature
coordinate. Task Rareness KDE uses the listed base-calibrated score family.
}
\label{tab:app_event_axis_score_families}
\\
\toprule
Task & Event descriptor & Task Rareness score family \\
\midrule
\endfirsthead

\toprule
Task & Event descriptor & Task Rareness score family \\
\midrule
\endhead

\bottomrule
\endfoot

Can
& Can-local fingertip contact geometry
& task-specific local feature percentile \\

Can
& contact end-effector pose
& task-specific local feature percentile \\

Can
& release end-effector pose
& task-specific local feature percentile \\

Lift
& box--end-effector contact point
& robust-distance percentile \\

Lift
& short-horizon box motion
& robust-distance percentile \\

Lift
& contact timing
& raw percentile \\

Square
& end-effector trajectory
& local feature kNN-DTM percentile \\

Square
& object--goal contact dynamics
& task-interest anchor percentile \\

Square
& peg-insert object motion
& task-interest anchor percentile \\

Transport
& bimanual pre-approach
& task-interest anchor percentile \\

Transport
& left gripper contact mode
& task-interest anchor percentile \\

Transport
& handoff geometry
& task-interest anchor percentile \\

ToolHang
& pre-hook approach
& task-interest anchor percentile \\

ToolHang
& grasp / hook contact
& task-interest anchor percentile \\

ToolHang
& gripper--tool contact mode
& task-interest anchor percentile \\

Push-T
& block--target contact dynamics
& task-interest anchor percentile \\

Push-T
& block target progress
& task-interest anchor percentile \\

Push-T
& late nearest keypoint
& task-interest anchor percentile \\

Franka Kitchen
& attempt order
& local feature kNN-DTM percentile \\

Franka Kitchen
& success order
& local feature kNN-DTM percentile \\

Block Pushing
& block2 motion trajectory
& local feature kNN-DTM percentile \\

Block Pushing
& agent--block2 contact mode
& task-interest anchor percentile \\

Block Pushing
& block2 contact path
& task-interest anchor percentile \\

\end{longtable}
\endgroup
\FloatBarrier

\subsection{Task-Feature Distribution-Shift Diagnostics}
\label{app:task_feature_kde_overlaps}

\begin{figure}[h]
    \centering
    \includegraphics[width=0.98\textwidth]{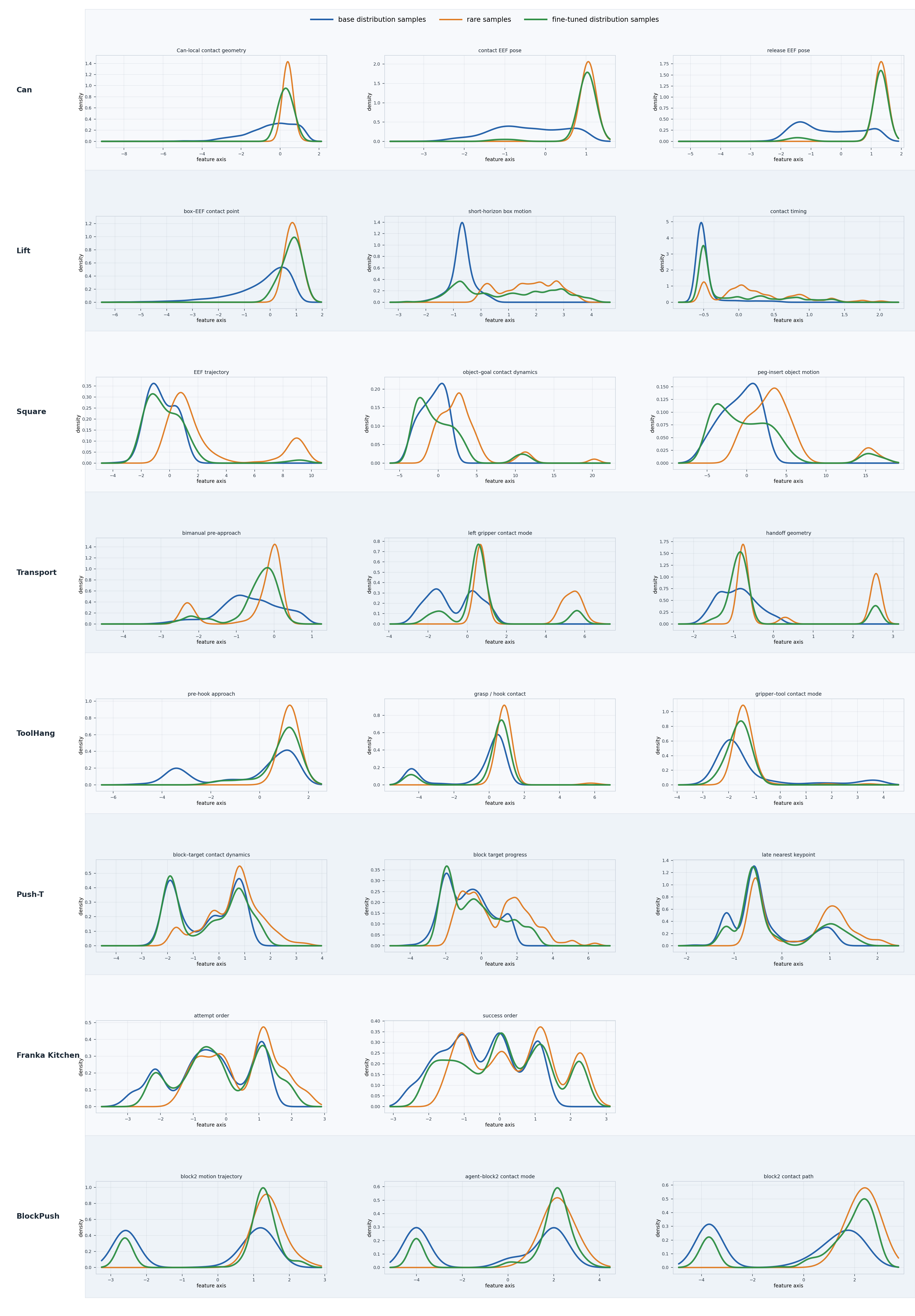}
    \caption{
    \textbf{Task-feature distribution-shift diagnostics across manipulation benchmarks.}
    Each panel shows a one-dimensional Task Feature KDE for a task-related event axis.
    Blue curves denote rollout samples from the base diffusion policy, orange curves denote rare-sampled rollout candidates, and green curves denote rollout samples from the fine-tuned policy after adaptation.
    The horizontal coordinate is the base-calibrated task-feature axis, oriented so that rightward movement corresponds to the rarer side of the base-policy rollout distribution.
    Densities are normalized within each panel and should be interpreted as distributional overlap and shift patterns rather than as directly comparable absolute density magnitudes across different event measures.
    Kitchen panels show Gaussian-smoothed KDEs of scalar embeddings of discrete attempt-order and success-order descriptors.
    }
    \label{fig:app_task_feature_kde_overlaps_all}
\end{figure}

Figure~\ref{fig:app_task_feature_kde_overlaps_all} overlays the base-policy,
adapted-policy, and rare-sampled rollout banks on the Task Feature KDE axes.
Rightward shifts correspond to movement toward the rare side of the
base-calibrated task-feature coordinate. Quantitative summaries are reported in
Tables~\ref{tab:app_task_feature_kde_components}
and~\ref{tab:app_task_level_task_feature_kde}.

\section{Baseline KDE Ablation Details}
\label{app:ablation_all_metrics}

This section reports the component metrics for GDNB, the external baselines, and the pretrained checkpoint in Table~\ref{tab:exp2_four_objective}. Task Feature KDE metrics are computed on
task-feature coordinates. Task Rareness KDE metrics are computed on
base-calibrated rareness scores. For \textbf{Kitchen}, attempt-order and success-order are discrete subtask-order descriptors whose scalar base-calibrated embeddings are smoothed for KDE plots and metrics; the true-success-sequence diagnostic is evaluated categorically by sequence frequencies.

\begin{table}[H]
\centering
\scriptsize
\caption{
\textbf{Task Feature KDE component summary.}
Entries are percentages. Outside-base denotes regions where base relative
density is below the cutoff.
}
\label{tab:app_task_feature_kde_components}
\setlength{\tabcolsep}{2.4pt}
\renewcommand{\arraystretch}{1.05}
\resizebox{\linewidth}{!}{
\begin{tabular}{@{}lrrrrrrrrr@{}}
\toprule
Method
& Base dens.
& Out. mass
& Out. var
& Out. even.
& Out. div.
& DV bal.
& Feat. ret.
& Feat. br.
& Feat. bal. \\
\midrule
\textbf{GDNB (Ours)}
& 84.0 & 15.0 & 10.5 & 74.3 & 21.0 & \textbf{30.8}
& 66.3 & 78.2 & 66.3 \\

DPPO
& 81.5 & 15.3 & 11.6 & 71.2 & 20.4 & 27.6
& 59.3 & 66.1 & 50.3 \\

SIME
& 83.7 & 14.9 & 9.1 & 69.7 & 17.4 & 26.2
& 64.5 & 88.9 & 66.0 \\

SOE
& 81.6 & 17.5 & 7.4 & 69.3 & 14.1 & 21.6
& 61.6 & 89.6 & 66.1 \\

MimicGen
& 82.7 & 16.1 & 8.5 & 70.1 & 16.6 & 24.6
& 62.9 & 85.8 & 64.0 \\

DSRL
& 80.2 & 18.2 & 8.6 & 71.2 & 15.9 & 23.8
& 61.1 & \textbf{93.4} & \textbf{67.9} \\

Pretrained checkpoint
& 78.5 & 18.3 & 9.0 & 69.7 & 16.4 & 24.3
& 61.3 & 91.4 & 65.6 \\
\bottomrule
\end{tabular}
}

\vspace{0.25em}
{\footnotesize
Out. = outside-base; DV = density--variance; Feat. = feature.
}
\end{table}

\begin{table}[H]
\centering
\scriptsize
\caption{
\textbf{Task-level Task Feature KDE summary.}
Each cell reports density--variance balanced / feature balanced, in percent.
}
\label{tab:app_task_level_task_feature_kde}
\setlength{\tabcolsep}{2.4pt}
\renewcommand{\arraystretch}{1.05}
\resizebox{\linewidth}{!}{
\begin{tabular}{@{}lcccccccc@{}}
\toprule
Method
& Can
& Lift
& Square
& Transport
& ToolHang
& Push-T
& Kitchen
& Block Pushing \\
\midrule
\textbf{GDNB (Ours)}
& 1.6 / 41.6
& 30.1 / 53.1
& 34.1 / 47.2
& 40.8 / 59.4
& 27.0 / 72.0
& 31.3 / 76.3
& 38.2 / 91.6
& 43.5 / 89.6 \\

DPPO
& 1.3 / 38.6
& 18.4 / 48.2
& 1.9 / 8.9
& 20.4 / 16.7
& 35.2 / 73.4
& 58.8 / 38.2
& 36.3 / 89.3
& 48.8 / 89.4 \\

SIME
& 1.1 / 31.5
& 21.4 / 58.9
& 21.4 / 41.2
& 18.5 / 53.2
& 30.5 / 69.2
& 26.2 / 90.8
& 46.7 / 89.8
& 43.7 / 93.5 \\

SOE
& 1.2 / 31.4
& 19.9 / 50.3
& 16.7 / 51.7
& 16.3 / 53.9
& 21.0 / 69.4
& 22.7 / 88.0
& 28.6 / 90.4
& 46.4 / 93.6 \\

MimicGen
& 1.2 / 31.4
& 20.4 / 52.3
& 22.2 / 51.4
& 17.1 / 54.7
& 26.2 / 67.3
& 24.9 / 82.3
& 36.0 / 81.1
& 49.0 / 91.9 \\

DSRL
& 1.2 / 31.6
& 18.5 / 52.0
& 10.0 / 66.8
& 16.5 / 54.1
& 31.9 / 69.0
& 22.7 / 87.9
& 39.8 / 88.7
& 49.5 / 92.9 \\

Pretrained checkpoint
& 1.3 / 31.4
& 17.7 / 51.3
& 29.2 / 53.4
& 12.6 / 52.5
& 17.5 / 66.5
& 23.8 / 88.6
& 43.0 / 90.3
& 49.7 / 90.9 \\
\bottomrule
\end{tabular}
}
\end{table}

\subsection{Ablation Baseline Adaptations}
\label{app:ablation_baselines}

This appendix describes how the baseline methods in Experiment~2 are adapted to the common benchmark-scale rare-behavior discovery protocol. The goal of these comparisons is not to claim a bit-for-bit reproduction of every original training stack, but to instantiate each method's core mechanism under the same reset distribution, rollout budget, environment reward/success signal, and post-hoc evaluation events used for our method. Unless otherwise stated, all methods use the same fixed-start runner, the same base diffusion-policy checkpoint or seed demonstrations, the same native task rewards and success checks, and the same final-policy rollout protocol. The task-related rare events in Table~\ref{tab:exp2_task_axes} are used only for post-hoc measurement; they are not used as optimization rewards, exploration objectives, or selection labels for any baseline.

\paragraph{DPPO.}
DPPO fine-tunes a diffusion policy with policy-gradient reinforcement learning~\citep{ren2025diffusion}. In our adaptation, the imitation-trained diffusion policy is used as the initial stochastic policy. We collect rollouts with the same fixed-start runner used by the other methods, compute native environment rewards and advantages, and update the diffusion model with a PPO-style objective. For diffusion actions, the policy likelihood term is implemented through a reverse-transition log-probability surrogate along the continuous-SDE denoising process, so that the update remains compatible with diffusion-policy action chunks. The reported DPPO policy is the final fine-tuned policy evaluated by the same rollout and post-hoc rare event metrics as all other methods. This adaptation preserves DPPO's central idea---native-reward policy-gradient improvement of the diffusion policy---but it does not add explicit novelty rewards or task rare event labels.

\paragraph{SIME.}
SIME targets policy self-improvement through modal-level exploration and data selection~\citep{jin2025sime}. We implement it as a modal-exploration wrapper around the base diffusion policy. At each decision step, the policy samples multiple candidate action chunks. These chunks are embedded into a mode representation using PCA features, a mode dictionary, and a history-based novelty score. The executed chunk is selected to favor underrepresented modes while remaining valid under the policy sampler. Rollouts are then filtered by task success or segment quality and used for subsequent self-improvement fine-tuning with the same rehearsal protocol used by the other data-augmentation baselines. This preserves SIME's mode-level exploration and data-selection principle, while adapting its mode representation to diffusion action chunks and the fixed-start benchmark runner.

\paragraph{SOE.}
SOE explores on a learned action manifold rather than by arbitrary action-space noise~\citep{jin2025soe}. In our adaptation, we fit a compact action-manifold model from base-policy action chunks and seed demonstrations. The main implementation uses a compact PCA-based action manifold for stable cross-task deployment, and the learned variant uses a conditional encoder--decoder to map between action chunks and latent coordinates. Exploration is performed by perturbing the latent coordinate and decoding the result back to an action chunk, which is then executed with the same fixed-start runner. The resulting rollouts are evaluated and, when used for policy improvement, inserted through the same training-data interface as the other methods. This preserves the SOE principle of manifold-constrained perturbation, but replaces the original task-specific latent training and human-guidance components with a benchmark-uniform action-manifold adaptation.

\paragraph{MimicGen.}
MimicGen is a demonstration-synthesis baseline rather than a policy-exploration baseline~\citep{mandlekar2023mimicgen}. We therefore adapt it at the data-generation level. Starting from the same seed demonstrations, we segment demonstrations into object-centric subtasks, transform the corresponding motion segments to new object or scene contexts, replay the adapted segments with the task controller, and retain only demonstrations that pass the environment success checks. The successful synthetic demonstrations are converted into the same diffusion-policy training format as the other accepted data and used to fine-tune the original policy. For fairness in the final comparison, we evaluate the fine-tuned policy using the same fixed-start rollout bank and the same post-hoc task-related rare events. This adaptation keeps MimicGen's structured demonstration-generation mechanism, but it does not treat MimicGen as an online action sampler.

\paragraph{DSRL.}
DSRL improves a diffusion policy by learning a steering policy in the latent-noise space while keeping the base diffusion-policy weights frozen~\citep{wagenmaker2025dsrl}. In our adaptation, the base policy is frozen and a lightweight RL policy outputs latent-noise or initial-noise modifications for the diffusion sampler. The steering policy is trained with native environment reward using PPO or SAC, depending on the task interface, and the final frozen-policy-plus-steering composite is evaluated on the same fixed-start rollouts as the other methods. When rollout artifacts are required for downstream analysis, we save the trajectories generated by the final steering policy in the same replay-compatible format used by the other baselines. This preserves DSRL's central frozen-policy steering semantics and avoids direct diffusion-weight fine-tuning.


\section{Continuation Across Rare-Mining Rounds}
\label{app:multiround_continuation}

We evaluate whether rare mining saturates after one augmentation round on three
tasks. Rare1 denotes the accepted rare set from the first round, and Rare2
denotes the accepted rare set obtained by applying the same sampler to the
first-round fine-tuned policy with unchanged hyperparameters.

For each task-related axis, we report four descriptive quantities. \textbf{Retention}
is the fraction of Rare2 samples inside the Rare1 5--95 percentile interval.
\textbf{Novelty} is the complementary mass outside that interval. \textbf{TV}
is the total-variation distance between the Rare1 and Rare2 one-dimensional KDEs
on the calibrated task-axis score:
\[
    \mathrm{TV}(\mathrm{Rare1},\mathrm{Rare2})
    =
    \frac{1}{2}
    \int_0^1
    \left|
        p_{\mathrm{Rare1}}(s)-p_{\mathrm{Rare2}}(s)
    \right|\,ds .
\]
High retention means Rare2 preserves the previous rare family, while high TV
means the Rare2 score distribution differs strongly from Rare1.

We also report two shifts relative to the first-round fine-tuned policy.
$\Delta_{\mathrm{out}}$ is the change in mass outside the fine-tuned 5--95
percentile interval:
\[
    \Delta_{\mathrm{out}}
    =
    \Pr_{\mathrm{Rare2}}[s\notin I^{(\mathrm{ft})}_{5\text{--}95}]
    -
    \Pr_{\mathrm{Rare1}}[s\notin I^{(\mathrm{ft})}_{5\text{--}95}] .
\]
$\Delta_{>q95}$ is the corresponding change in mass above the fine-tuned 95th
percentile. Positive values indicate that Rare2 moves farther into the
fine-tuned tail; negative values indicate a more conservative next rare set.

Finally, \textbf{Spread ratio} is the Rare2/Rare1 ratio of median pairwise RMS
distance within each rare set, averaged over task axes. It is a descriptive
internal-diversity measure: values above one indicate that Rare2 is more spread
out than Rare1 on average, but this should be read together with retention and
outwardness.

\begin{table}[H]
\centering
\small
\caption{
\textbf{Consecutive-round rare-mining diagnostic.}
Retention, Novelty, TV, $\Delta_{\mathrm{out}}$, $\Delta_{>q95}$, and Spread
ratio are averaged over task-related axes. TV measures the KDE distance between
Rare1 and Rare2 on calibrated axis scores. Spread ratio is the Rare2/Rare1
median pairwise RMS distance within the rare set.
}
\label{tab:app_multiround_summary}
\setlength{\tabcolsep}{3.5pt}
\renewcommand{\arraystretch}{1.05}
\begin{tabular*}{\textwidth}{@{\extracolsep{\fill}}lccccccc@{}}
\toprule
Task
& 1st/2nd round success rate
& Ret.
& Nov.
& TV
& $\Delta_{\mathrm{out}}$
& $\Delta_{>q95}$
& Spread ratio \\
\midrule

ToolHang
& 0.630/0.682
& 0.838
& 0.162
& 0.348
& $-0.052$
& $+0.049$
& 1.304 \\

Transport
& 0.855/0.859
& 0.936
& 0.064
& 0.332
& $-0.202$
& $-0.033$
& 1.354 \\

Can
& 0.947/0.998
& 0.005
& 0.995
& 0.957
& $+0.637$
& $+0.856$
& 7.265 \\
\bottomrule
\end{tabular*}
\end{table}

\paragraph{Interpretation.}
Can shows the clearest continuation signal. Rare2 is almost entirely outside
Rare1 support, has the largest TV distance, moves strongly into the fine-tuned
tail, and has much larger internal spread. This indicates a new contact/release
pose family rather than saturation of the first rare family.

ToolHang shows rare-family retention with mild expansion. Rare2 preserves most
of Rare1 support, but still adds nonzero novelty and a small positive high-tail
shift. Its main change is approach/contact-orientation variation within the
same broad ToolHang rare family.

Transport should be read conservatively. Rare2 is distributionally different
from Rare1, but it has high retention, low novelty, negative outward shift
relative to the fine-tuned policy, and only moderate spread increase. This is
best interpreted as local handoff/contact geometry reweighting and rare-support
consolidation, not broad new-mode discovery.

\end{document}